\documentclass{article}

\PassOptionsToPackage{numbers,compress}{natbib}

\usepackage[nonatbib,final]{neurips_2022}
\usepackage[ref,caption]{leaf}
\usepackage{algorithm}
\usepackage{algpseudocode}
\usepackage{pifont}

\newcommand{\specialcell}[2][c]{
\begin{tabular}[#1]{@{}c@{}}#2\end{tabular}}

\title{PAC-Bayes Compression Bounds So Tight That They Can Explain Generalization}

\author{
  Sanae Lotfi\thanks{Equal contribution.} \qquad
  Marc Finzi$^*$ \qquad
  Sanyam Kapoor$^*$ \qquad
  Andres Potapczynski$^*$ \\ \\
  \textbf{Micah Goldblum} \qquad
  \textbf{Andrew Gordon Wilson} \\ \\
  New York University
}

\begin{document}

\maketitle

\begin{abstract}
While there has been progress in developing non-vacuous generalization bounds for deep neural networks, these bounds tend to be uninformative about why deep learning works.
In this paper, we develop a compression approach based on quantizing neural network parameters in a linear subspace, profoundly improving on previous results to provide state-of-the-art generalization bounds on a variety of tasks, including transfer learning.
We use these tight bounds to better understand the role of model size, equivariance, and the implicit biases of optimization, for generalization in deep learning.
Notably, we find large models can be compressed to a much greater extent than previously known, encapsulating Occam’s razor. We also argue for
data-independent bounds in explaining generalization.
\end{abstract}

\section{Introduction}

Despite many more parameters than the number of training datapoints, deep learning models generalize extremely well and can even fit random labels \citep{zhang2021understanding}. These observations are not explained through classical statistical learning theory such as VC-dimension or Rademacher complexity which focus on uniform convergence over the hypothesis class \citep{nagarajan2019uniform}. The PAC-Bayes framework, by contrast, provides a convenient way of constructing generalization bounds where the generalization gap depends on the deep learning model found by training rather than the hypothesis set as a whole. Using this framework, many different potential explanations have been proposed drawing on properties of a deep learning model that are induced by the training dataset, such as low spectral norm \citep{neyshabur2017pac}, noise stability   \citep{arora2018stronger}, flat minima  \citep{hochreiter1997flat}, derandomization \citep{negrea2020defense}, robustness, and compression \citep{arora2018stronger,zhou2019nonvacuous}.

In this work, we show that neural networks,  when paired with structured training datasets, are substantially more compressible than previously known.
Constructing tighter generalization bounds than have been previously achieved, we show that this compression \emph{alone} is sufficient to explain many generalization properties of neural networks.

In particular:
\begin{enumerate}
    \item We develop a new approach for training compressed neural networks that adapt the compressed size to the difficulty of the problem. We train in a random linear subspace of the parameters \citep{Li2018MeasuringTI} and perform learned quantization. Consequently, we achieve extremely low compressed sizes for neural networks at a given accuracy level, which is essential for our tight bounds. (See \cref{sec:our-method}).

    \item Using a prior encoding Occam's razor and our compression scheme, we construct the best generalization bounds to date on image datasets, both with data-dependent and data-independent priors. We also show how transfer learning improves compression and thus our generalization bounds, explaining the practical performance benefits of pre-training. (See \cref{sec:empirical-non-vacuous-bounds}).

    \item PAC-Bayes bounds only constrain the adaptation of the prior to the posterior. For bounds constructed with data-dependent priors, we show that the prior alone achieves performance comparable to the generalization bound. Therefore we argue that bounds constructed from data-independent priors are more informative for understanding generalization. (See \Cref{sec:data-dep-priors}).

    \item Through the lens of compressibility, we are able to help explain why deep learning models generalize on structured datasets like CIFAR-10, but not when structure is broken such as by shuffling the pixels or shuffling the labels. Similarly, we describe the benefits of equivariant models, e.g. why CNNs outperform MLPs. Finally, we investigate double descent and whether the implicit regularization of SGD is necessary for generalization. (See \cref{sec:understanding-gen}).
\end{enumerate}

\begin{figure}[!ht]
\centering
    \begin{tabular}{ccc}
    \hspace{-1cm}\includegraphics[width=.28\linewidth]{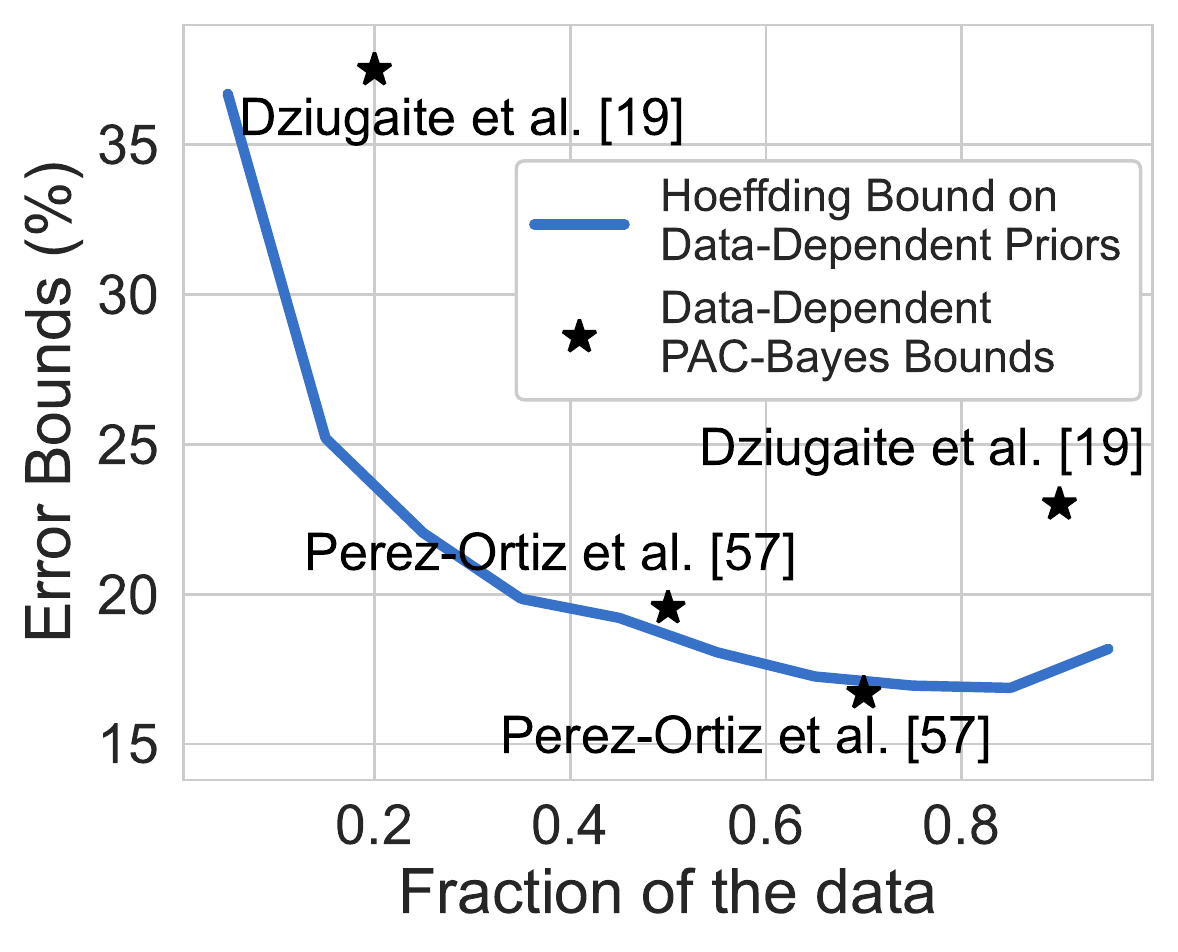} &
    \hspace{-0.6cm}\includegraphics[width=.34\linewidth]{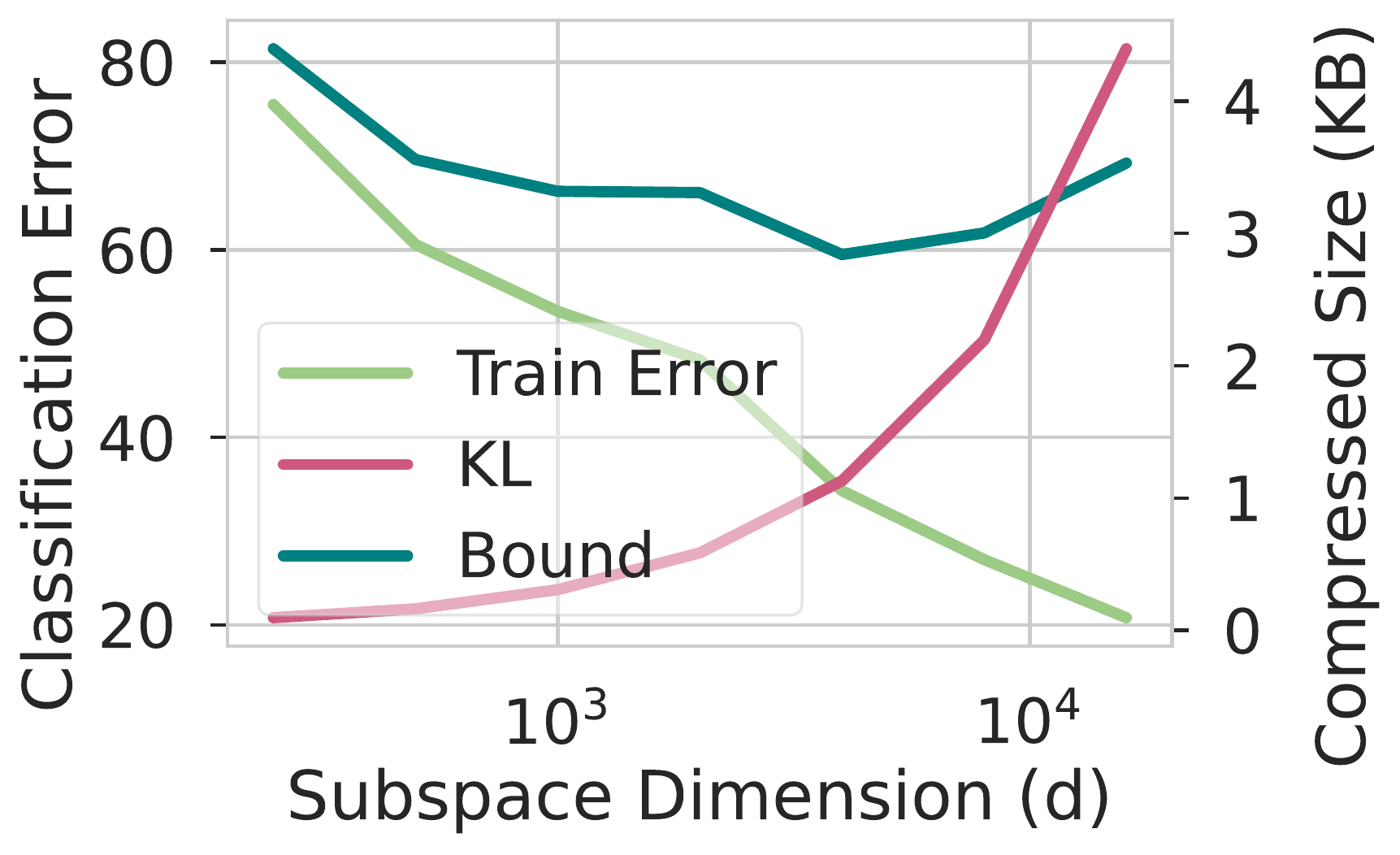} &
    \hspace{-0.2cm}\includegraphics[width=.3\linewidth]{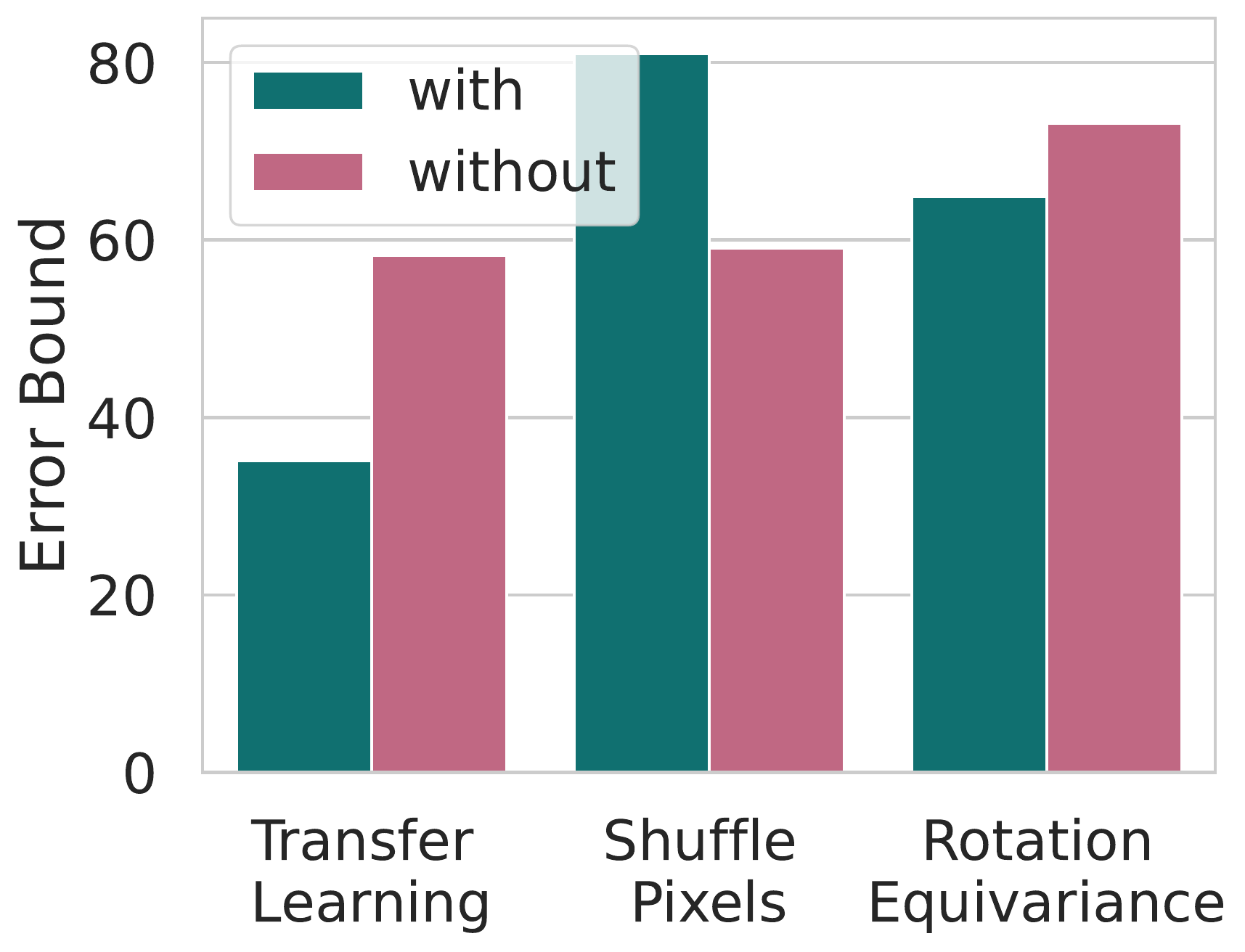}
      \\
    \hspace{-0.2cm}{\small (a) Pitfalls of data-dependent bounds} &
    \hspace{-0.3cm}{\small (b) Adaptive compression} &
    \hspace{-0.4cm}{\small (c) Deep learning phenomena}
    \end{tabular}
   \caption{
   \textbf{The power of data-independent subspace compression bounds for explaining deep learning phenomena.} Bounds for CIFAR-10 except (c)-rotation, which is rotMNIST.
   \textbf{(a)} We show that the simple Hoeffding bound computed \emph{only} on the data-dependent prior and evaluated on the remainder of the training data (essentially measuring validation loss) achieves error bounds that are competitive or even better than data-dependent bounds obtained by previous works, showing that data-dependent PAC-Bayes bounds do not explain generalization any further than the prior alone. Instead, data-independent bounds are more informative for understanding generalization (see \cref{sec:data-dep-priors}).
   \textbf{(b)} Training error, the KL term (compressed model size measured in KB), and our PAC-Bayes bound as the subspace dimension is varied. For a fixed network, our method provides an adaptive compression scheme that trades off compressed size with training error, finding the optimal bound for a given model and dataset.
   \textbf{(c)} 
   We compute our data-independent bounds for model trained \emph{with} and \emph{without}: transfer learning, shuffling the pixels, and the rotation-equivariance property.  Our bounds identify the positive impact of transfer learning, how breaking structure in the data by shuffling pixels hurts the model, and that rotationally equivariant models improve generalization on rotated data. Each of these interventions impact the compressibility of the models.
   See Section~\ref{sec:understanding-gen} for more details.
   }
    \label{fig:conceptual}
\end{figure}

We emphasize that while we achieve state-of-the-art results in both data-dependent bounds and data-independent bounds through our newly developed compression approach, our goal is to leverage these tighter bounds to understand generalization in neural networks. Among others, \cref{fig:conceptual} highlights some of our observations regarding (a) data-dependent bounds, (b) how our method trades-off between data fit and model compression in relation to generalization, and (c) the explanation of several deep learning phenomena through model compressibility using our bounds.

All code to reproduce results is available \href{https://github.com/activatedgeek/tight-pac-bayes}{here}.

\section{Related Work}
\label{sec:related-work}

\noindent \textbf{Optimizing the PAC-Bayes Bound.} \quad
\citet{dziugaite2017computing} obtained the first non-vacuous generalization bounds for deep stochastic neural networks on binary MNIST. The authors constructed a relaxation of the \citet{langford2001bounds} bound and optimized it to find a posterior distribution that covers a large volume of low-loss solutions around a local minimum obtained using SGD. \citet{rivasplata2019pac} further extended the idea by developing novel relaxations of PAC-Bayes bounds based on \citet{blundell2015weight}.

\noindent \textbf{Model Compression and PAC-Bayes Bounds.} \quad
Noting the robustness of neural networks to small  perturbations \citep{hinton1993keeping, hochreiter1997flat, langford2001not, langford2002quantitatively, keskar2016large, neyshabur2017pac, chaudhari2019entropy},
\citet{arora2018stronger} developed a compression-based approach that uses noise stability.
Additionally, they used the ability to reconstruct weight matrices with random projections to study generalization of neural networks.
Subsequently, \citet{zhou2019nonvacuous} developed a PAC-Bayes bound that uses the representation of a compressed model in bits, and added noise stability through the use of Gaussian posteriors and Gaussian mixture priors. Furthermore, they achieved even smaller model representations through pruning and quantization \citep{han2016deep,cheng2018model}.
Our compression framing is similar to \citet{zhou2019nonvacuous} but with key improvements. First, we train in a lower dimensional subspace using intrinsic dimensionality \citep{Li2018MeasuringTI} and FiLM subspaces \citep{perez2018film} which proves to be more effective and adaptable than pruning. Second, we develop a more aggressive quantization scheme with variable length code and quantization aware training. Finally, we exploit the increased compression provided by transfer learning and data-dependent priors.

\noindent \textbf{Data-Dependent Priors.} \quad
\citet{dziugaite2021data} demonstrated that for linear PAC-Bayes bounds such as \citet{thiemann2017strongly}, a tighter bound can be achieved by
choosing the prior distribution to be data-dependent, i.e., the prior is trained to concentrate around low loss regions on held-out data.
More precisely, the authors show that the optimal data-dependent prior is the conditional expectation of the posterior given a subset of the training data. They approximate this data-dependent prior by solving a variational problem over Gaussian distributions.
They evaluate the bounds for SGD-trained networks on data-dependent priors obtaining tight bounds on MNIST, Fashion MNIST, and CIFAR-10.
In a similar vein,
\citet{perez2021tighter} combine data-dependent priors \citep{dziugaite2021data} with the PAC-Bayes with Backprop (PBB) \citep{rivasplata2019pac} to obtain \textit{state-of-the-art} PAC-Bayes non-vacuous bounds for MNIST and CIFAR-10 using data-dependent priors.

\noindent \textbf{Downstream Transferability.} \quad
\citet{ding2022pactran} investigate different correlates of generalization derived from PAC-Bayes bounds to predict the transferability of various upstream models; however, because of this different aim they do not actually compute the full bounds.

Our focus is to achieve better bounds in order to better understand generalization in deep neural networks. For example, we investigate the effects of transfer learning, equivariance, and stochastic training on the bounds, and argue for the importance of data-independent bounds in explaining generalization. We summarize improvements of our bounds relative to prior results in \cref{tab:multicol}.

\begin{table}[!t]
\caption{\textbf{Non-vacuous PAC-Bayes bounds obtained on popular image classification datasets in deep learning.} $\star$ indicates bounds obtained using data-dependent priors (\Cref{sec:data-dep-priors}).
\xmark $\,$ indicates that either the method does not support multi-class problems or that it is completely reliant on data-dependent priors and therefore cannot result in data-independent bounds. Additionally, we add Binary MNIST for reference to a benchmark used in earlier works.
}
\begin{center}
\begin{adjustbox}{width=\linewidth}
\begin{tabular}{cccccccc}
    \hline
    &\multicolumn{5}{c}{Non-vacuous PAC-Bayes bounds ($\%$)}\\
    \cline{2-7}
    Reference & Binary MNIST & MNIST & FMNIST & CIFAR-10 & CIFAR-100 & ImageNet
    \\
    \hline
    \citet{dziugaite2017computing}& $16.1$ & \xmark & \xmark & \xmark
    & \xmark & \xmark
    \\
    \citet{rivasplata2019pac}& $2.2$ & \xmark & \xmark & \xmark
    & \xmark & \xmark
    \\
    \citet{zhou2019nonvacuous} &  & $46$ & 91.6 & 100 & 100 & 96.5
    \\
     \citet{dziugaite2021data} &  & $11^\star$ & $38^\star$ & $23^{\star}$ & \xmark & \xmark
    \\
    \citet{perez2021tighter} &  & $21.7 / 1.5^\star$ & $49.1$ & $90.0/16.7^\star$ & $100$ & \xmark
    \\
    Our bounds &  & $\mathbf{11.6}$/$\mathbf{1.4}^\star$ & $\mathbf{32.8}$/$\mathbf{10.1}^\star$ & $\mathbf{58.2}$/$ \mathbf{16.6}^\star$ & $\mathbf{94.6}$/$\mathbf{44.4}^\star$ & $\mathbf{93.5}$/$\mathbf{40.9}^\star$ \\
    \hline
\end{tabular}
\end{adjustbox}
\end{center}
\label{tab:multicol}
\end{table}

\section{A Primer on PAC-Bayes Bounds} \label{sec:pac}

PAC-Bayes bounds are fundamentally an expression of Occam's razor: simpler descriptions of the data generalize better.
As an illustration, consider the classical generalization bound on a finite hypothesis class. Let $\hat{R}(h) = \frac{1}{n} \sum_{i=1}^{n} \ell\left(h\left(x_{i}\right), y_{i}\right)$ be the empirical risk of a hypothesis $h \in \mathcal{H}$, with $|\mathcal{H}| < \infty$. Let $\ell$ be the $0$-$1$ loss, and let $R(h) = \mathbb{E}[\hat{R}(h)]$ denote the population risk.
With probability at least $1-\delta$, the population risk of hypothesis $h$ using $n$ data samples satisfies
\begin{equation}
    R(h) \le \hat{R}(h) + \sqrt{\frac{\logdet{\mathcal{H}} +\log(1/\delta)}{2 n}}.
\end{equation}
In other words, the population risk is bounded by the empirical risk and a complexity term $\logdet{\mathcal{H}}$ which counts the number of bits needed to specify any hypothesis $h \in \mathcal{H}$.

But what if we don't believe that each hypothesis is equally likely? If we consider a prior distribution over the hypothesis class that concentrates around likely hypotheses, then we can construct a variable length code that uses fewer bits to specify those hypotheses. Note that if
$P$ is a prior distribution over $\mathcal{H}$, then any given hypothesis
$h$ will take $\log_2\frac{1}{P(h)}$ bits to represent when using an optimal compression code for $P$.
This prior may result in a smaller complexity term as long as the hypotheses that are consistent with the data are also likely under the prior, regardless of the size of the hypothesis class.

Moreover, the number of bits required can be reduced from $\log_2\frac{1}{P(h)}$ to  ${\mathbb{KL}(Q \parallel P)}$ by considering a distribution of ``good'' solutions $Q$. If we don't care which element of $Q$ we arrive at, we can gain some bits \emph{back} from this insensitivity (which could be used to code a separate message). The average number of bits to code a sample from $Q$ using the prior $P$ is the cross entropy $\mathbb{H}(Q,P)$ and we get $\mathbb{H}(Q)$ bits back from being agnostic about which sample $h\sim Q$ to use, yielding the KL-divergence between $Q$ and $P$: ${\mathbb{H}(Q,P)-\mathbb{H}(Q) = \mathbb{KL}(Q \parallel P)}$.

With these improvements on the finite hypothesis bound ---
replacing $\logdet{\mathcal{H}}$ with $\mathbb{KL}(Q \parallel P)$, and sampling a hypothesis $h \in \mathcal{H}$ ---
we arrive (with minor bookkeeping) at the PAC-Bayes bound introduced in \citet{mcallester1999pac}. This last bound states that with probability at least $1-\delta$,
\begin{equation} \label{eq:mac}
    \underset{h \sim Q}{\mathbb{E}}\left[R\left(h\right)\right]
    \le
    \underset{h \sim Q}{\mathbb{E}}[\hat{R}\left(h\right)]+ \sqrt{\frac{\mathbb{KL}(Q \parallel P) + \log(n/\delta)+ 2}{2n - 1}}.
\end{equation}

Many refinements of \cref{eq:mac} have been developed
\citep{langford2001bounds, maurer2004pac, catoni2007pacbayes, thiemann2017strongly} but retain the same character. That is, the lower the ratio of the KL-divergence to the number of data points $n$, the lower the gap between empirical and expected risk. In this work, we use the tighter \citet{catoni2007pacbayes} variant of the PAC-Bayes bound (see \cref{sec:catoni} for details).

\textbf{Universal Prior.} \quad
Leveraging Occam's razor, we can define a prior that explicitly penalizes the minimum compressed length of the hypothesis, also known as the universal prior \citep{solomonoff1964formal}: $P(h) = 2^{-K(h)}/Z$, where $K$ is the \emph{prefix} Kolmogorov complexity \citep{hutter2008algorithmic} of $h$ (the length of the shortest program that produces $h$ and also delimits itself), and $Z \le 1$.\footnote{The universal prior similar to the discrete hypothesis prior from \citet{zhou2019nonvacuous} but setting $m(h) = 2^{-2\log_2 l(h)}$ rather than the flat $m(h) = 2^{-72}$.} Using a point mass posterior on a single hypothesis $h^*$, we get the following upper-bound

\begin{equation*}
    \mathbb{KL}\left(\mbf{1}_{[h=h^*]} \parallel P(h)\right)=\log \tfrac{1}{P(h^*)} \le K(h^*)\log 2 \le l(h^*)\log 2 + 2\log l(h^*),
\end{equation*}

where $l(h)$ is the length of a given program that reproduces $h$ not including the delimiter. For convenience, we can condition on using the same method for compression and decompression for all elements of the prior.
Lastly, we can improve the tightness of the previous PAC-Bayes bound by reducing
the compressed length $l(h^{*})$ of the hypothesis $h^*$ that we found during training.

\textbf{Model Compression.}\quad
\emph{Model compression} aims to find nearly equivalent models that can be expressed in fewer bits either for deploying them in mobile devices or for improving their inference time on specialized hardware \citep{cheng2017survey,cheng2018model}. For computing PAC-Bayes generalization bounds, however, we only care about the model size. Therefore, we can employ compression methods which may otherwise be unfavorable in practice due to worse computational requirements. \emph{Pruning} and \emph{quantization} are among the most widely used methods for model compression. In this work, we rely on quantization (\cref{subsec:quant}) to achieving tighter generalization bounds.

\section{Tighter Generalization Bounds via Adaptive Subspace Compression}
\label{sec:our-method}

Training a neural network involves taking many gradient steps in a high-dimensional space $\reals^D$. Although $D$ may be large, the loss landscape has been found to be simpler than typically believed \citep{Dauphin2014IdentifyingAA,Goodfellow2015QualitativelyCN,Garipov2018LossSM}. Analogous to the notion of intrinsic dimensionality more generally, \citet{Li2018MeasuringTI} searched for the lowest dimensional subspace in which the network can be trained and still fit the training data. The weights of a neural network $\theta\in\reals^D$ are parametrized in terms of an initialization $\theta_0$ and a projection $w\in \reals^d$ to a lower dimensional subspace through a fixed matrix $P \in \reals^{D\times d}$,
\begin{align}
    \theta = \theta_0 + P w. \label{eq:ID}
\end{align}
To facilitate favorable conditioning during optimization, $P$ is chosen to be approximately orthonormal ${P^\top P \approx I_{d\times d}}$. For scalability, \citet{Li2018MeasuringTI} use random normal matrices of the form ${P \sim \gaussian{0, 1}^{D\times d} / \sqrt{D}}$, and their sparse approximations \citep{Li2006VerySR,Le2013FastfoodAK}.

In its original form, intrinsic dimensionality (ID) is only used as a scientific tool to measure the complexity of the learning task. Unlike methods like pruning, intrinsic dimension \citep{Li2018MeasuringTI} scales with complexity of the task --- more complex tasks require a larger intrinsic dimension.
Subsequently, we find that ID combined with quantization can serve as an effective model compression method.
We note that ideas similar to the intrinsic dimensionality of a model have been explored in the context of model compression for estimating bounds  \citep{arora2018stronger}.
For our work, the ability to find the intrinsic dimension $d \ll D$ has profound implications for the compressibility of the models, and therefore our ability to construct generalization bounds. As demonstrated by \citet{zhou2019nonvacuous}, the compressibility of a neural network has a direct connection to generalization and allows us to compute non-vacuous PAC-Bayes bounds for transfer learning.

Resting upon ID, our key building blocks to achieve tight generalization bounds are composed of (i) a new scalable method to train an intrinsic dimensionality neural network parameterized by \cref{eq:ID} (\cref{subsec:better_id}), and (ii) a new approach to simultaneously train both the quantized neural network weights and the quantization levels for maximum compression (\cref{subsec:quant}). Our complete method is summarized in \cref{alg:pac}.

\subsection{Finding Better Random Subspaces} \label{subsec:better_id}

To further improve upon the scalability and effectiveness of the projections $P$ used by \citet{Li2018MeasuringTI}, we introduce three novel projector constructions.

\begin{figure}[!t]
\centering
    \begin{tabular}{ccc}
    \includegraphics[width=.29\linewidth]{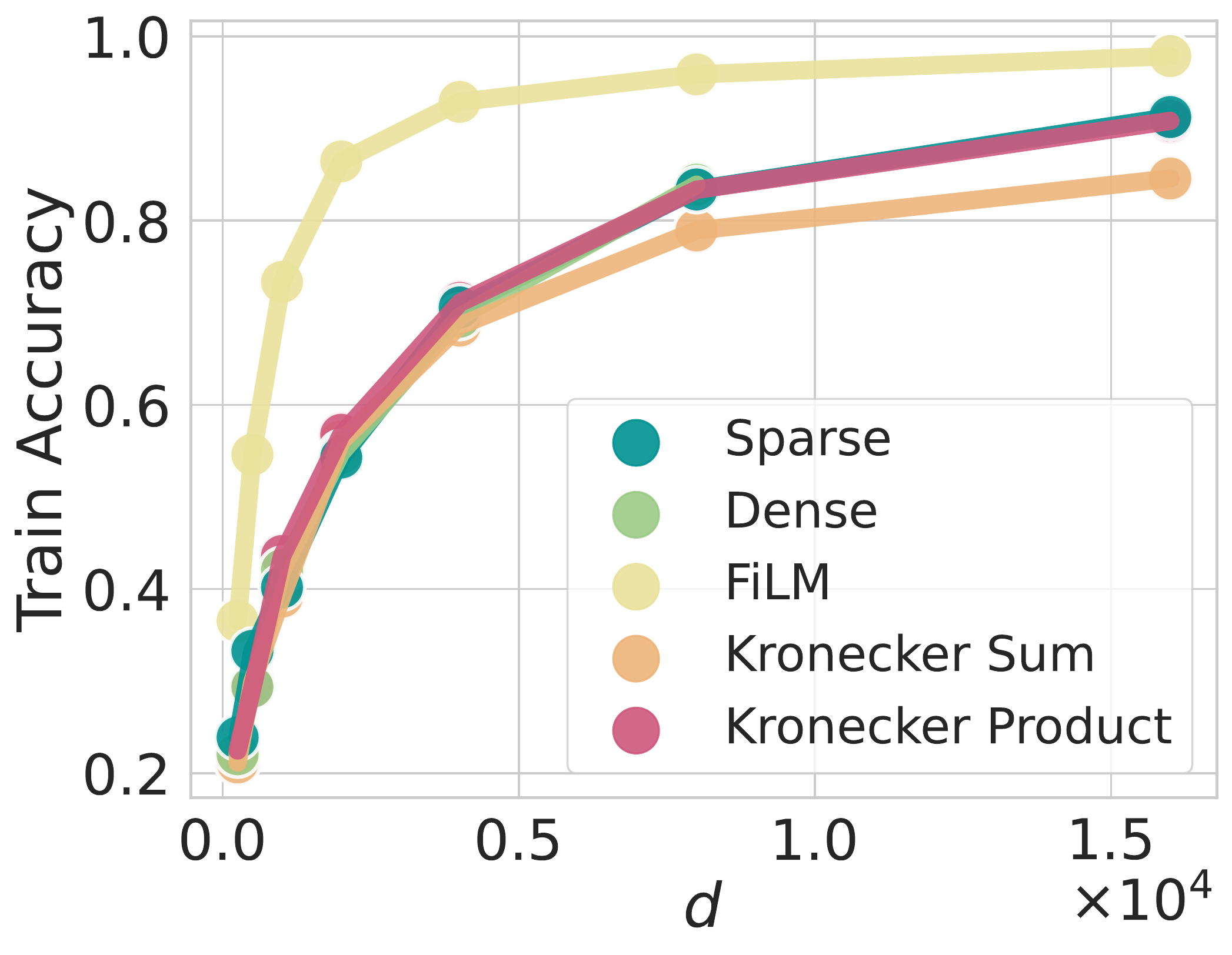} &
    \includegraphics[width=.29\linewidth]{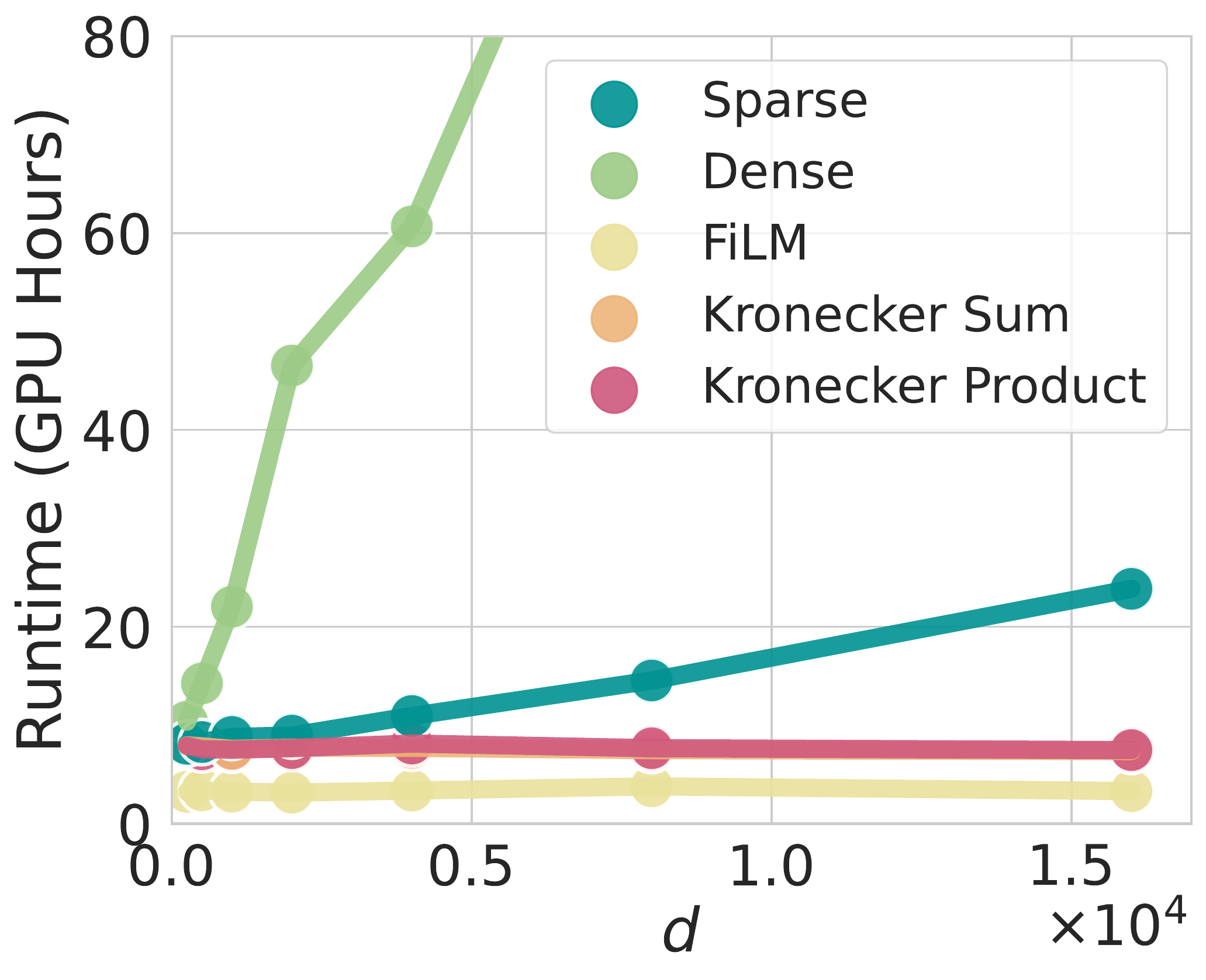} &
    \includegraphics[width=.29\linewidth]{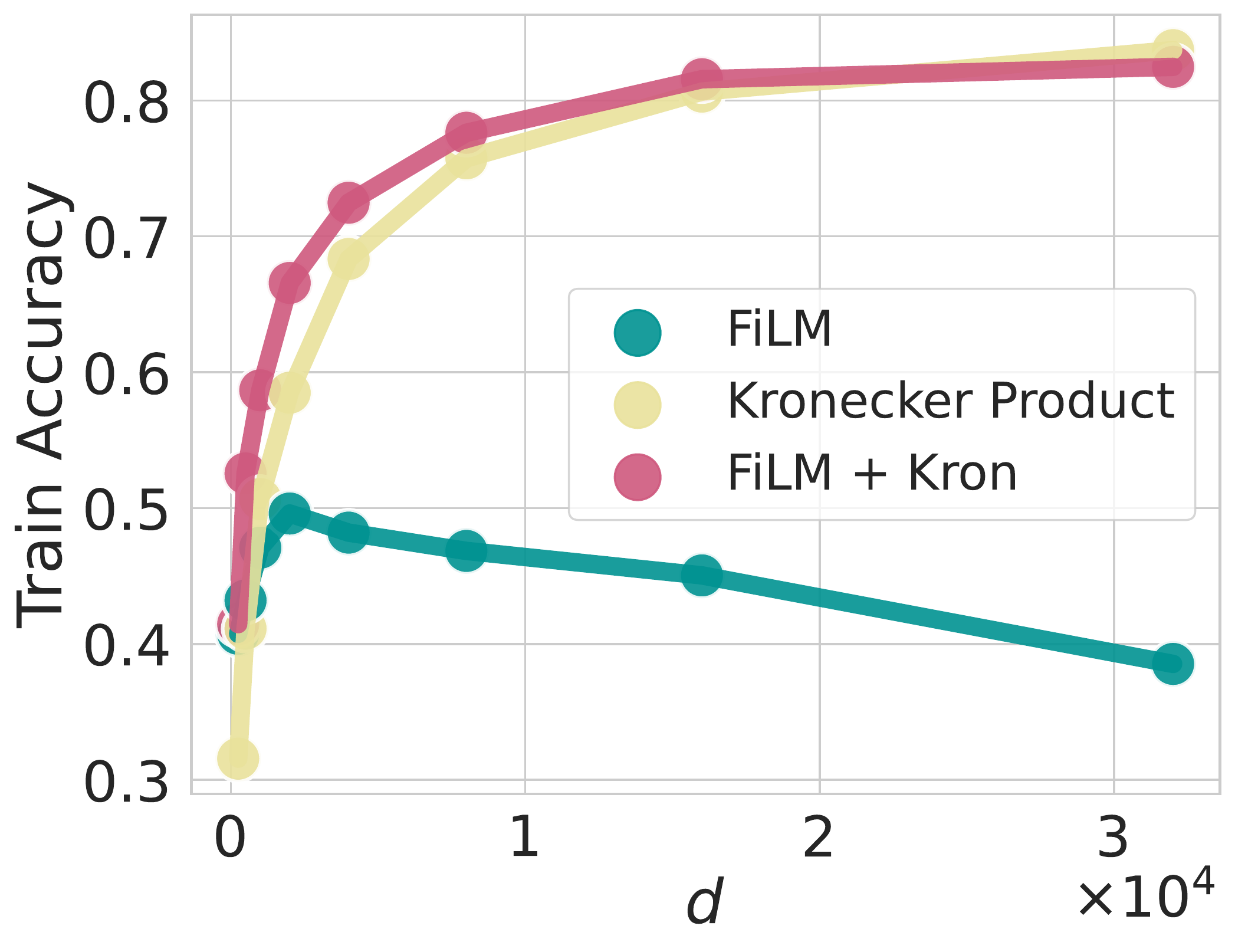}
    \end{tabular}
   \caption{\textbf{Effective and scalable projection operators.}
   \textbf{(Left)} Different projection operators $P$ (\cref{subsec:better_id}) used for transfer learning from Imagenet to CIFAR-10 on a ResNet-34 across different subspace dimensions $d$. Kronecker product, Sparse, and Dense perform almost identically \textbf{(Center)} Kronecker product runs with substantially reduced the runtime cost compared to the Sparse or Fastfood matrices used by \citet{Li2018MeasuringTI}.
   \textbf{(Right)} Training from scratch on CIFAR-10. The FiLM projector alone is unable to fit the data when training from scratch, and instead a sum of FiLM and Kronecker Product projectors perform the best.}
    \label{fig:projectors_comp}
\end{figure}

\textbf{Kronecker Sum Projector.} \quad
Using the Kronecker product $\otimes$, we construct the matrix ${P_\oplus = (\mathbf{1}\otimes R_1 + R_2 \otimes \mathbf{1})/\sqrt{2D}}$ where $R_1,R_2\sim \mathcal{N}(0,1)^{\sqrt{D}\times d}$ and $\mathbf{1}$ is the vector of all ones in $\reals^{\sqrt{D}}$.
Noting that $R_1 \ci R_2$ and that the entries are standard normal,  $P^\top_\oplus P_\oplus=I_{d\times d}+\mathcal{O}(1/\sqrt{D})$.

\textbf{Kronecker Product Projector.} \quad
Alternatively, we form the matrix $P_\otimes=Q_1\otimes Q_2/\sqrt{D}$ with the smaller $Q_1,Q_2\sim \mathcal{N}(0,1)^{\sqrt{D}\times \sqrt{d}}$, and again this matrix is approximately orthogonal: $P^\top_\otimes P_\otimes=(Q_1^\top Q_1/\sqrt{D})\otimes(Q_2^\top Q_2/\sqrt{D}) =I\otimes I+O(D^{-1/4}) = I_{d\times d}+O(D^{-1/4})$. \footnote{As neither $D$ nor $d$ is typically a perfect square, we concatenate a dense random matrix to pad out the difference between $D$, $d$, and a perfect square. As $(\sqrt{D}+1)^2=D+2\sqrt{D}+1$, we have that the size of this padding is at most $\sqrt{D}\times \sqrt{d}$, so it does not increase the asymptotic cost of performing the matrix vector multiplies.}

The matrix vector multiply $w \mapsto Pw$ for both of the above projectors can be performed in time $O(d\sqrt{D})$ and $O(\sqrt{dD})$ respectively, rather than the $O(dD)$ that is required by the dense random matrix. \cref{fig:projectors_comp} demonstrates the runtime speedup and training performance improvement in comparison to the methods using by \citet{Li2018MeasuringTI}. Notably, the Kronecker-structured projections retain the fidelity of the dense random matrix while being orders of magnitude faster than the alternative operators when scaling to larger values of $d$. In other words, the Kronecker-structured projectors are as good as the dense projectors for generating random linear subspaces of a given size, but are much scalable.

\textbf{FiLM projector.} \quad
BatchNorm parameters have an disproportionate effect on the downstream task performance relative to their size. This observation has been used in Featurewise independent Linear Modulation (FiLM) \citep{perez2018film,dumoulin2018feature} for efficient control of neural networks in many different settings. Several authors have explored performing fine-tuning for transfer learning solely on these parameters and the final linear layer \citep{kanavati2021partial}. Drawing on these observations, we construct a projection matrix $P_\mathrm{FiLM}$ where only columns corresponding with BatchNorm or head parameters are non-zero and sampled from $\mathcal{N}(0,1)^d/\sqrt{D}$, which we also show in \cref{fig:projectors_comp}. While the FiLM projector is highly effective for transfer learning (shown in \cref{fig:projectors_comp} left), the performance saturates quickly when training from scratch. For this reason, when training from scratch we employ the sum $P_{\mathrm{FiLM}+\otimes} = (P_\mathrm{FiLM}+P_\otimes)/\sqrt{2}$, which outperforms the two projectors individually as shown in \cref{fig:projectors_comp} right.

\subsection{Quantization Scheme and Training} \label{subsec:quant}

Through quantization, the average number of bits used per parameter can be substantially reduced. When optimizing purely for model size rather than efficiency on specialized hardware, we can choose non-linearly spaced quantization levels which are learned, and use variable length coding schemes as shown in \citet{han2016deep}. Additionally, the straight through estimator has been central to learning weights in binary neural networks \citep{hubara2016binarized}. We combine these ideas to simultaneously optimize the quantized weights and the quantization levels for maximum compression.

Given the full precision weights ${w = [w_1,\dots,w_d] \in \reals^d}$ and a vector ${c = [c_1,\dots c_L] \in \reals^L}$ of $L$ quantization levels, we construct the quantized vector ${\hat{w} = [\hat{w}_1,\dots,\hat{w}_d]}$ such that ${\hat{w}_i =c_{q(i)}}$ where ${q(i)= \argmin_k |w_i-c_k|}$. The quantization levels $c$ are learned alongside $w$, where the gradients are defined using the straight through estimator \citep{bengio2013estimating, yin2019ste}:
\begin{align}
    \frac{\partial \hat{w}_i}{\partial w_j} = \delta_{ij} \quad
      \text{and}
      \quad
      \frac{\partial \hat{w}_i}{\partial c_k} = 1_{[q(i)=k]}
\end{align}
We initialize $c$ with uniform spacing between the minimum and maximum values in parameter vector $w$ or k-means \citep{choi2016towards}.
To further compress the network, we use a variable length code in the form of arithmetic coding \citep{mackay2003information}, which takes advantage of the fact that certain quantization levels are more likely than others. Given probabilities $p_k$ (empirical fractions) for cluster $c_k$, arithmetic coding of $w$ takes at most
$\ceil{d \times \mathbb{H}(p)} + 2$ bits, where $\mathbb{H}(p)$ is the entropy $\mathbb{H}(p) = -\sum_k p_k \log_2 p_k$. For a small number of quantization levels, arithmetic coding yields better compression than Huffman coding.

In summary, we use $\ceil{d \times \mathbb{H}(p)}+2$ bits for coding the quantized weights $\hat{w}$, $16L$ bits for the codebook $c$ (represented in half precision), and additional $L \times \ceil{\log_2{d}}$ bits for representing the probabilities $p_k$, arriving at $l(w) \le \ceil{d \times \mathbb{H}(p)}+ L \times (16+ \ceil{\log_2{d}}) +2$. As we show in \cref{subsec:prior}, we optimize over the subspace dimension $d$ and the number of quantization levels $L$ and any other hyperparameters, by including them in the compressed description of our model, contributing only a few extra bits.

\subsection{Transfer Learning}
For transfer learning, we replace $\theta_0$ with a learned initialization $\theta_{\dset}$ that is found using the pretraining task and data $\dset$. With the ID compression, the universal prior $P(h\mid \theta_\dset) \propto 2^{-K(h\mid \theta_\dset)}$ will place higher likelihood on solutions $\theta$ that are close to the pre-training solution $\theta_{\dset}$.

\begin{small}
\begin{algorithm}[!t]
  \caption{
    Compute PAC-Bayes Bound.
  }\label{alg:pac}
\begin{algorithmic}[1]
\State \textbf{Inputs:} Neural network $f_{\theta}$, Training dataset $\left\{x_{i}, y_{i}\right\}_{i=1}^{n}$, Clusters $L$, Intrinsic dimension $d$, Confidence $1-\delta$, and Prior distribution $P$.
\vspace{0.03cm}
\Function{\texttt{COMPUTE\_BOUND}}{$f_{\theta}, L, d, \left(x_{i}, y_{i}\right)_{i=1}^{n}, \delta, P$}
    \State $w$ $\leftarrow$ \texttt{TRAIN\_ID}($f_{\theta}$, $d$, $\left(x_{i}, y_{i}\right)_{i=1}^{n}$) \Comment{(\cref{subsec:better_id})}
    \State $\hat{w}$ $\leftarrow$ \texttt{TRAIN\_QUANTIZE}($w$, $L$, $\left(x_{i}, y_{i}\right)_{i=1}^{n}$)
    \State Compute quantized train error $\hat{R}\left(\hat{w}\right)$.
    \State $\mathbb{KL}\left(Q,P\right) \leftarrow$ \texttt{GET\_KL}$\left(\hat{w}, P\right)$ \Comment{(\cref{sec:pac})}
    \State \textbf{return} \texttt{GET\_CATONI\_BOUND}($\hat{R}\left(\hat{w}\right)$, $\mathbb{KL}\left(Q, P\right)$, $\delta$, $n$) \Comment{(\cref{sec:pac})}
\EndFunction
\vspace{0.03cm}
\Function{\texttt{TRAIN\_QUANTIZE}}{$w,L, \left(x_{i}, y_{i}\right)_{i=1}^{n}$} \Comment{(\cref{subsec:quant})}
    \State Initialize $c \leftarrow$ \texttt{GET\_CLUSTERS}($w, L$)
    \For{$i=1$ to \texttt{quant\_epochs}}
    \State $c \leftarrow c -\rho \nabla_{c}\mathcal{L}\left(w, c\right)$ and
           $w \leftarrow w - \rho \nabla_{w}\mathcal{L}\left(w, c\right)$
    \EndFor
    \State \textbf{return} $\hat{w}$
\EndFunction
\vspace{0.03cm}
\Function{\texttt{GET\_KL}}{$\left(\hat{w}, P\right)$}
    \State $c$, \texttt{count} $\leftarrow$ \texttt{GET\_UNIQUE\_VALS\_COUNTS}($\hat{w}$)
    \State \texttt{message\_size} $\leftarrow$ \texttt{DO\_ARITHMETIC\_ENCODING}($\hat{w}$, $c$, \texttt{count})
    \State \texttt{message\_size} $\leftarrow$ \texttt{message\_size} + \texttt{hyperparam\_search} \Comment{(\cref{subsec:prior})}
     \State \textbf{return} $\texttt{message\_size} + 2\times\log\left(\texttt{message\_size}\right)$
\EndFunction
\end{algorithmic}
\end{algorithm}
\end{small}

\section{Empirical Non-Vacuous Bounds}
\label{sec:empirical-non-vacuous-bounds}
Combining the training in structured random subspaces with our choice of learned quantization, we produce extremely compressed but high performing models. Using the universal prior, we bound the generalization error of these models and optimize over the degree of compression via the subspace dimension and other hyperparameters as summarized in \cref{alg:pac}. 
We additionally describe hyperparameters, architecture specifications for each experiment, and other experimental details in \Cref{app-sec:bound-hypers-optim}.
In the following subsections, we apply our method to generate strong generalization bounds in the data-independent, data-dependent, and transfer learning settings.

\subsection{Non-Vacuous PAC-Bayes Bounds}
\label{sec:non-vacuous-bounds-exp}

We present our bounds for the \textit{data-independent} prior
in \cref{table:our-bounds}.
We derive the first non-vacuous bounds on FashionMNIST, CIFAR-10, and CIFAR-100 without data-dependent priors.
These results have particular significance, as we argue in \cref{sec:data-dep-priors} that using data-dependent priors are not explanatory about the learning process.
In particular, we improve over the compression bound results obtained by~\citet{zhou2019nonvacuous} on MNIST from $46\%$ to $11.55\%$ and on ImageNet from $96.5\%$ to $94.1\%$.
In terms of compression,
we dramatically improve the rates as we reduce the compressed size for the best MNIST bound by $94\%$ bringing it down from $6.23$ KB to $0.38$KB with LeNet5 and, on ImageNet, by
$87\%$ bringing it down from $358$ KB to $46.3$ KB with MobileViT. Since we perform transfer learning with an ImageNet-trained checkpoint, we omit transfer learning experiments on the ImageNet (downstream) dataset.
The tightness of our SOTA subspace compression bounds allows us to improve the understanding of several deep learning phenomena as discussed in \cref{sec:understanding-gen}. See \cref{app-sec:model_training} for model architectures and \cref{app-sec:additional results} for additional results.

\begin{table}[!t]
  \caption{\textbf{Our PAC-Bayesian subspace compression bounds
  compared to \emph{state-of-the-art} (SOTA) bounds.} All results are with $95\%$ confidence, i.e. $\delta = .05$.
  The sign $\dagger$ refers to data-independent SOTA numbers that we computed using \citep{perez2021tighter}, which we run on the additional datasets. \\}
  \label{table:our-bounds}
  \centering
\begin{adjustbox}{width=.9\linewidth}
\begin{tabular}{l|cccc}
\hline
 Dataset & \multicolumn{2}{c}{Data-independent priors} & \multicolumn{2}{c}{Data-dependent priors} \\
 \cline{2-5}
 & Err. Bound (\%) & SOTA (\%) & Err. Bound  (\%) & SOTA (\%) \\
  \hline
 MNIST  & $\mathbf{11.6}$ & $21.7$ \citep{perez2021tighter}   & $\mathbf{1.4}$ & $1.5$ \citep{perez2021tighter} \\
 \phantom{+} + SVHN Transfer & $\mathbf{9.0}$ & $16.1^\dagger$ & & \\
 \hline
 FashionMNIST  & $\mathbf{32.8}$ & $46.5^\dagger$ & $\mathbf{10.1}$ & $38$ \citep{dziugaite2021data}\\
 \phantom{+} + CIFAR-10 Transfer & $\mathbf{28.2}$ & $30.1^\dagger$ & & \\
 \hline
 CIFAR-10 & $\mathbf{58.2}$ & $89.9^\dagger$ & $ \mathbf{16.6}$ & $16.7$ \citep{perez2021tighter}\\
 \phantom{+} + ImageNet Transfer& $\mathbf{35.1}$& $54.2^\dagger$ &  &  \\
 \hline
 CIFAR-100  & $\mathbf{94.6}$ & $100^\dagger$ & $\mathbf{44.4}$ & --\\
 \phantom{+} + ImageNet Transfer &$\mathbf{81.3}$ & $98.1^\dagger$ &   &  \\
 \hline
 ImageNet  & $\mathbf{93.5}$ & $96.5$ \citep{zhou2019nonvacuous}  & $\mathbf{40.9}$ & --\\
 \hline
\end{tabular}
\end{adjustbox}
\end{table}

\subsection{Data-Dependent PAC-Bayes Bounds}
\label{sec:data-dep-priors}

So far, we demonstrated the strength of our bounds on \emph{data-independent} priors, where we considerably improve on the state-of-the-art. However, a number of recent papers have considered data-dependent priors as a way of achieving tighter bounds \citep{perez2021tighter, dziugaite2021data}. In this setup, the training data $\mathcal{D}=\{(x_i,y_i)\}_{i=1}^n$ is partitioned into two parts, $\mathcal{D}_a$ and $\mathcal{D}_b$, with length $n-m$ and $m$. The first dataset is used to construct a data-dependent prior $P(h\mid\mathcal{D}_a)$, and then the bound is formed over the remaining part of the process: the adaptation of the prior $P(h\mid\mathcal{D}_a)$ to the posterior $Q(h)$ using the data $\mathcal{D}_b$. The empirical risk is computed over $\mathcal{D}_b$ only.
Intuitively, using dataset $\mathcal{D}_a$ it is possible to construct a much tighter prior over the possible neural network solutions.

In our setting, similar to transfer learning, we use the prior $P_{\mathcal{D}_a}(\theta) = 2^{-K(\theta|\theta_{\mathcal{D}_a})}/Z$ where for compression we use $\theta = \theta_{\mathcal{D}_a} + Pw$, and $\theta_{\mathcal{D}_a}$ is the solution found by training the model (without random projections) on the data ${\mathcal{D}_a}$ rather than initializing randomly. With these data-dependent priors, we achieve the best bounds in \cref{table:our-bounds}.

However, our adaptive approach exposes a significant downside of data-dependent priors. To the extent that PAC-Bayes bounds can be used for explanation, data-dependent bounds only provide insights into the procedure used to adapt the prior $P_{\mathcal{D}_a}(\theta)$ to the posterior $Q$ using $\mathcal{D}_b$: any learning that is done in finding $P_{\mathcal{D}_a}(\theta)$ is not constrained or explained by the bound (we have no information on the compressibility of $\theta_{\mathcal{D}_a}$). Given the ability to adapt the size of the KL to the difficulty of the problem, it is possible to squeeze all of the learning into $P_{\mathcal{D}_a}(\theta)$ and none in this adaption to $Q$.
 This phenomenon happens as the $\mathbb{KL} \rightarrow 0$, which we find happens empirically (or very nearly so) across splits of the data, and especially when $n-m$ is large. Setting $Q(\theta)=\mathbf{1}_{[\theta=\theta_{\mathcal{D}_a}]}$, the KL has only the contribution from the optimization over $d$: $\mathbb{KL}(Q||P_{\mathcal{D}_a}) \le \log D$. We find that the bound is nothing more than a variant of the simple Hoeffding bound where $\mathcal{D}_b$ is the validation set
$R\left(\theta_{\mathcal{D}_a}\right)
    \le
    \hat{R}_{\mathcal{D}_b}\left(\theta_{\mathcal{D}_a}\right) + \sqrt{\frac{\log(Dm/\delta)+ 2}{2m - 1}}$.

We can see this phenomenon in  \cref{fig:conceptual}(a) where we compare existing data-dependent bounds to the simple Hoeffding bound applied directly to the data-dependent prior which was trained on only a small fraction of the data.  We can consider the Hoeffding bound as the simplest data-dependent bound without any fine-tuning so that the \textit{prior}, a single pre-trained checkpoint, is directly evaluated on held-out validation data with no KL-divergence term. If another data-dependent bound cannot achieve significantly stronger guarantees than the prior Hoeffding bound, then it only explains that neural networks generalize because the priors already have low validation error which is no explanation for generalization at all.  Indeed, we see in Figure~\ref{fig:conceptual} that the strength of existing data-dependent bounds relies almost entirely on the a priori properties of the data-dependent prior rather than constraining the actual learning process through compressibility. Similarly, from a minimum description length (MDL) perspective, data-independent bounds can be used to provide a lossless compression of the training data, whereas data-dependent bounds cannot (see \cref{sec:modelsize}).

We also note that with data-dependent priors, optimization over the subspace dimension selects very low dimensionality, even if the data does not have low intrinsic dimension. Because most of the data fitting is moved into fitting the prior, the bound selects a low complexity solution with respect to the prior without hurting data fit by choosing a low subspace dimensionality (\cref{app-sec:ddp}).

By contrast, data-independent bounds explain generalization for the entirety of the learning process.
Similarly, our transfer learning bounds meaningfully constrain what happens in the fine-tuning on the downstream task, but they do not constrain the prior determined from the upstream task.

\subsection{Non-Vacuous PAC-Bayes Bounds for Transfer Learning}
\label{sec:non-vacuous-bounds-transfer-exp}

By directly interpreting PAC-Bayes bounds through the lens of compression, we immediately see the benefits of using an upstream dataset for transfer learning.
Transfer learning allows us to constrain the prior $P(\theta \mid \theta_{\mathcal{D}_a})$ around parameters consistent with the upstream dataset $\mathcal{D}_a$, reducing the KL-divergence between the prior and the posterior and leading to even tighter bounds as we show in \cref{table:our-bounds}.
Our tighter data-independent transfer learning bounds
provide a theoretical certification that transfer learning can improve generalization.
Our PAC-Bayes transfer learning approach also indicates that transfer learning can boost generalization whenever codings optimized on a pre-training task are more efficient for encoding a downstream posterior than an a priori guess made before seeing data.  By contrast, downstream tasks which greatly differ
from the upstream task may only be consistent with models that are not compressible under the learned prior, a scenario that describes negative transfer. See \cref{app-sec:transfer-learning-bounds} for more details.

\section{Understanding Generalization through PAC-Bayes Bounds} \label{sec:understanding-gen}

The classical viewpoint of uniform convergence focuses on properties of the hypothesis class as a whole, such as its size.  In contrast, PAC-Bayes shows that the ability to generalize is not merely a result of the hypothesis class but also a result of the particular dataset and the characteristics of the individual functions in the hypothesis class.
After all, many elements of our hypothesis class are not compressible, yet in order to guarantee generalization, we choose the ones that are.  Real datasets actually contain a tremendous amount of structure, or else we could not learn from them as famously argued by \citet{Hume1978} and No Free Lunch theorems \citep{wolpert1997no,giraud2005toward}.  This high degree of structure in real-world datasets is reflected in the compressibility of the functions (i.e. neural networks) we find in our hypothesis class which fit them.  

In this section, we examine exactly how dataset structure manifests in compressible models by applying our generalization bounds, and we see what happens when this structure is broken, for example by shuffling pixels or fitting random labels.  Corrupting the dataset degrades both  compressibility and generalization.

\begin{figure}[!ht]
    \centering
    \includegraphics[width=.7\linewidth]{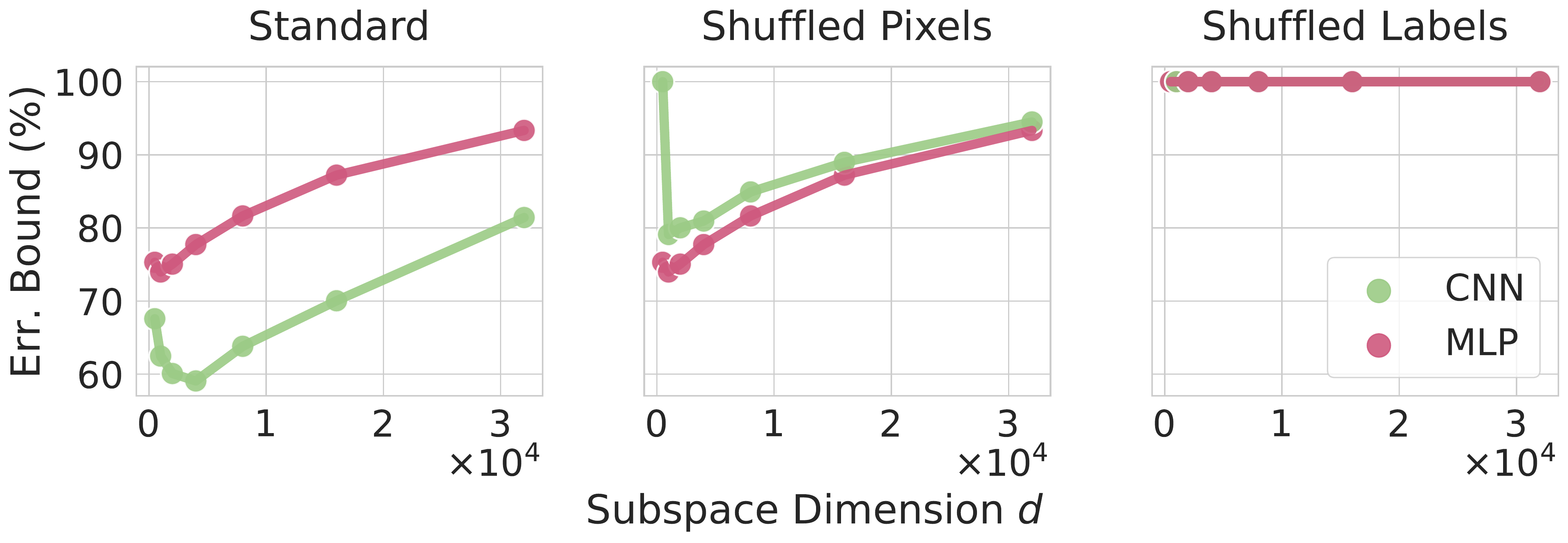}
    \caption{
    \textbf{Breaking structure in the data and the model}.
    Our PAC-Bayes bound computed using various subspace dimensions for a fixed size CNN and MLP, both with $500$k parameters. We train on \textbf{(left)} CIFAR-10, \textbf{(center)} CIFAR-10 with shuffled pixels, \textbf{(right)} CIFAR-10 with shuffled labels. Structure in the dataset induces structure in the model. As structure is removed from the dataset, models which fit the data become much less compressible, hence generalize worse.
    }
    \label{fig:fitting_random_labels}
\end{figure}

\textbf{MLPs vs CNNs.} \quad
It is well known that convolutional neural networks (CNNs) generalize much better than standard multilayer perceptrons (MLPs) with alternating fully-connected layers and activation functions on image classification problems, even when controlling for the number of parameters.
In terms of our generalization bounds, this is reflected in the improved compressibility of CNNs when compared to MLPs.
In \cref{fig:fitting_random_labels} (left), we see on CIFAR-10 that a CNN finds a more structured and lower description length explanation of the training data than an MLP with the same number of parameters, therefore achieving better generalization bounds. See \cref{sec-app:breaking_structure} for experimental details.

\textbf{Shuffled Pixels.} \quad
However, when the image structure is broken by shuffling the pixels, we find that CNNs are no better at generalizing than MLPs. For this dataset, CNNs become substantially less compressible and hence our bounds show them generalizing worse than MLPs, see \cref{fig:fitting_random_labels} (center).
MLPs do not suffer when this structure is broken since they never used it in the first place.

\textbf{Shuffled Labels.} \quad
When the structure of the dataset is entirely broken by shuffling the labels, the compressibility of the models (both for CNNs and MLPs) which fit the random labels is lost. Regardless of the subspace dimension used, our generalization bounds are all at $100\%$ error as shown in \cref{fig:fitting_random_labels} (right). It is not possible to fit the training data using low subspace dimensions, and when using a large enough dimension to fit the data, the compressed size of the model is larger than the training data and hence the generalization bounds are vacuous.

\textbf{Equivariance.} \quad
Designing models which are \emph{equivariant} to certain symmetry transformations has been a guiding principle for the development of data-efficient neural networks in many domains \citep{cohen2016group,cohen2018spherical,thomas2018tensor,weiler2019general,finzi2020generalizing,jumper2021highly}. While intuitively it is clear that respecting dataset symmetries severely improves generalization, relatively little has been proven for neural networks \citep{lyle2020benefits, elesedy2021provably,zhu2021understanding,bietti2021sample,elesedy2022group}.
We compress and evaluate rotationally equivariant ($C_8$) and non-equivariant Wide ResNets \citep{weiler2019general,zagoruyko2016wide} trained on MNIST and a rotated version of MNIST. As shown in \cref{fig:equivariance}, the rotationally equivariant models are more compressible and provably generalize better than their non-equivariant counterparts when paired with a dataset that also has the rotational symmetry. See \cref{sec:app_equivariance} for further details.

\textbf{Is Stochasticity Necessary for Generalization?} \quad
It is widely hypothesized that the implicit biases of SGD help to find solutions which generalize better. For example, \citet{arora2018optimization} argue that there is no regularizer that replicates the benefits of gradient noise. \citet{wu2020noisy}, \citet{smith2020generalization}, and \citet{li2021validity} advocate that gradient noise is necessary to achieve state-of-the-art performance.
In comparison, recent work by \citet{geiping2022stochastic} shows that full-batch gradient descent can match state-of-the-art performance, and \citet{izmailov2021bayesian} shows that full-batch Hamiltonian Monte Carlo sampling generalizes significantly better than mini-batch MCMC and stochastic optimization.

We train ResNet-18 and LeNet5 models on CIFAR-10 and MNIST, respectively, using full-batch and stochastic gradient descent with different intrinsic dimensionalities.
We provide the training details in \Cref{app-sec:stochasticity}, but unlike \citet{geiping2022stochastic}, we do not use flatness-seeking regularization for full-batch training.
For MNIST with LeNet5, the best generalization bounds that we obtain are $11.55\%$ and $11.20\%$ using stochastic gradient descent (SGD) and full-batch training respectively. The best generalization bounds that we obtain for CIFAR-10 with ResNet-18 are $74.68\%$ and $75.3\%$ using SGD and full-batch training respectively.
We also extend this analysis to SVHN to MNIST transfer learning with LeNet5 and obtain PAC-Bayes bounds of $9.0\%$ and $8.7\%$ using SGD and full-batch training respectively.

These close theoretical guarantees on the generalization error for both SGD and full-batch training suggest that while the implicit biases of SGD may be helpful, they are not at all necessary for understanding why neural networks generalize well.
We expand this analysis in \Cref{app-sec:stochasticity}.

\textbf{Double Descent.} \quad
Our bounds are also tight enough to predict the double descent phenomenon noted in \citet{Nakkiran2020DeepDD}.  See \cref{app-sec:double-descent} for a depiction of these experiments and a discussion of their significance.

\section{Discussion}
\label{sec:discussion}
In this work, we constructed a new method for compressing deep learning models that is highly adaptive to the model and to the training dataset. Following Occam's prior, which considers shorter compressed length models to be more likely, we construct state-of-the-art generalization bounds across a variety of settings. Through our compression bounds, we show how generalization relates to the structure in the dataset and the structure in the model, and we are able to explain aspects of neural network generalization for natural image datasets, shuffled pixels, shuffled labels, equivariant models, and stochastic training.

\textbf{Limitations.} \quad
Despite the power of our compression scheme and the ability of our bounds to faithfully describe the generalization properties of many modeling decisions and phenomena, we are scratching the surface of explaining generalization.
Our compression bounds prefer models with a smaller number of parameters as shown in \cref{sec:modelsize}, instead of larger models which actually tend to generalize better. While we achieved better model compression than previous works, it is unlikely that we are close to theoretical limits. Maybe through nonlinear parameter compression schemes we might find that larger deep learning models are more compressible than smaller models.
There is also a question of whether uncompressed models can be related and bounded by the performance of the compressed models, perhaps leveraging ideas from \citet{nagarajan2019deterministic} and others investigating this question.
Additionally, while our bounds show that the compressibility of our models implies generalization, we make no claims about the reverse direction.
However, we believe that model compression and Occam's razor have yet untapped explanatory power in deep learning.

\subsection*{Acknowledgements}

We thank Pavel Izmailov and Nate Gruver for helpful discussions.
This research is supported by NSF CAREER IIS-2145492, NSF I-DISRE 193471, NIH R01DA048764-01A1, NSF IIS-1910266, NSF 1922658 NRT-HDR, Meta Core Data Science, Google AI Research, BigHat Biosciences, Capital One, and an Amazon Research Award. This work is also supported in part through the NYU IT High Performance Computing resources, services, and staff expertise.

\bibliographystyle{plainnat}
\bibliography{totalbib}

\clearpage
\appendix

\vbox{%
\hsize\textwidth
\linewidth\hsize
\vskip 0.1in
\hrule height 4pt%\p@
  \vskip 0.25in
  \vskip -\parskip%
\centering
{\LARGE\bf
Supplementary Material for \\
PAC-Bayes Compression Bounds So Tight \\
That They Can Explain Generalization
\par}
\vskip 0.29in
  \vskip -\parskip
  \hrule height 1pt
  \vskip 0.09in%
}

\section*{Appendix Outline}
The appendix is organized as follows. 
\begin{itemize}
\item In \cref{app-sec:additional results}, we report results for additional bounds for SVHN and ImageNet. We also report the compression size corresponding to our best bound values and compare it to the compression size obtained through standard pruning. Furthermore, in \cref{sec:corrupted_incompressible} we prove why models cannot both be compressible and fit random labels.
\item In \cref{subsec:prior}, we describe how optimization over hyperparameters like the intrinsic dimension impact the PAC-Bayes bound
\item In \cref{app-sec:transfer-learning-bounds}, we show how our PAC-Bayes bound benefit from transfer learning.
\item In \cref{app-sec:ddp}, we discuss data-dependent priors and their effect on the subspace dimension optimization.
\item In \cref{app-sec:bound-hypers-optim}, we detail our experimental setup including models, datasets, and hyperparameter settings for training and bound computation.
\item In \cref{sec:app_equivariance}, we provide a compression perspective to why equivariant models may be more desirable for generalization.
\item In \cref{app-sec:stochasticity}, we further discuss how our 
through our PAC-Bayes compression bounds, we provide evidence that SGD is not necessary for generalization.
\item In \cref{sec:modelsize}, we ablate the model size and show how it impacts our bounds and compressibility, we identify the best performing size of models for our bounds.
\item In \cref{app-sec:double-descent}, we present our observations on double descent and their preditability from our PAC-Bayes bounds.
\item In \cref{sec:catoni}, we expand our theoretical discussion and emphasize conceptual differences between our method and previous ones in the literature.
\item Lastly, in \cref{sec:licences} we provide licensing information on the datasets we use.
\end{itemize}

\section{Additional Results}
\label{app-sec:additional results}

In addition to the results reported in \cref{table:our-bounds}, we report the best bounds for SVHN and ImageNet-1k as well as the corresponding compressed size in \cref{table:additional-di-our-bounds,table:additional-dd-our-bounds}. 
In \cref{table:additional-di-our-bounds} we show how compressing the model via intrinsic dimension (ID) yields better results than standard pruning.
In this table, we basically run our method but substitute ID with pruning and then proceed by quantizing the remaining weights and encoding 
them through arithmetic encoding. When pruning we used the standard iterative procedure following \citet{han2016deep}, for the MNIST model
we pruned 98.8\% of the weights, for the FMNIST model 97.0\% of the weights, for the SVHN model 98.8\% of the weights and
for both the CIFAR-10 and CIFAR-100 models we pruned 52.1\% of the weights and stopped there as the accuracy dropped significantly 
if we kept pruning.

\textbf{Error bars on our bounds:} We re-run the bounds computation for $10$ times and observe that the values are consistent. On average, we obtain $\pm 0.5\%$ variation in our bounds for models trained from scratch and $\pm 0.1\%$ variation for transfer learning models.

\begin{table}[!ht]
  \caption{Using our subspace method rather than pruning yields substantially higher compression ratios and hence tighter generalization bounds.
  We report our error bounds ($\%$) and compressed size ($\mathbb{KL}$ (KB)), $1$ KB = $8192$ bits. First, we compress the model weights using ID, quantizing its values and then storing them through arithmetic encoding. We then report
  the bounds obtained by only switching ID to standard pruning. All results are data-independent and obtained with $95\%$ confidence, i.e. $\delta = .05$.}
  \label{table:additional-di-our-bounds}
  \centering
\begin{adjustbox}{width=0.75\linewidth}
\begin{tabular}{l|cccc}
\hline
 Dataset & \multicolumn{2}{c}{ID + Quant + Arith} & \multicolumn{2}{c}{Pruning + Quant + Arith} \\
 \cline{2-5}
 & Err. Bound (\%) &  $\mathbb{KL}$ (KB) & Err. Bound (\%) & 
 $\mathbb{KL}$ (KB)\\
  \hline 
 MNIST  & 
 $\mathbf{11.6}$ & $0.4$  & 47.9 & 6.5\\
 \phantom{+} + SVHN Transfer & 
 $\mathbf{9.0}$ & $0.4$\\
 \hline
 FashionMNIST  & 
 $\mathbf{32.8}$ & $0.8$ & 54.9 & 3.5 \\
 \phantom{+} + CIFAR-10 Transfer & 
 $\mathbf{28.2}$ & $0.9$ \\
 \hline
  SVHN & 
  $\mathbf{36.1}$ & $1.3$ & 74.4 & 4.3\\
 \phantom{+} + ImageNet Transfer & 
 $\mathbf{29.1}$ & $1.4$ \\
 \hline
 CIFAR-10 & 
 $\mathbf{58.2}$ & $1.2$ & 100.0 & 57.8 \\
 \phantom{+} + ImageNet Transfer& 
 $\mathbf{35.1}$& $1.0$ \\
 \hline
 CIFAR-100  & 
 $\mathbf{94.6}$ & $4.1$ & 99.9 & 50.7 \\
 \phantom{+} + ImageNet Transfer & 
 $\mathbf{81.3}$ & 2.8  \\
 \hline
\end{tabular}
\end{adjustbox}
\end{table}

\begin{table}[!ht]
  \caption{Our PAC-Bayesian Subspace Compression Bounds
  with data-dependent priors compared to state-of-the-art PAC-Bayes non-vacous data-dependent bounds. All results are obtained with $95\%$ confidence, i.e. $\delta = .05$. 
  }
  \label{table:additional-dd-our-bounds}
  \centering
\begin{adjustbox}{width=0.5\linewidth}
\begin{tabular}{l|cc}
\hline
 Dataset & Err. Bound (\%) & SoTA (\%) \\
  \hline 
 MNIST  &  $\mathbf{1.4}$ & $1.5$ \citep{perez2021tighter} \\
 \hline
 FashionMNIST & $\mathbf{10.1}$ & $38$ \citep{dziugaite2021data}\\
 \hline
  SVHN & $\mathbf{8.7}$ & -- \\
 \hline
 CIFAR-10 & $ \mathbf{16.6}$ & $16.7$ \citep{perez2021tighter}\\
 \hline
 CIFAR-100  &  $\mathbf{44.4}$ & --\\
 \hline
  ImageNet  & $\mathbf{40.9}$ & -- \\
 \hline
\end{tabular}
\end{adjustbox}
\end{table}

\subsection{Models that can fit random labels cannot be compressed} \label{sec:corrupted_incompressible}

Our ability to construct nonvacuous generalization bounds rests on the ability to construct models which both fit the training data and are highly compressible. However, when the structure in the dataset has been completely destroyed by shuffling the labels, then we do not find that our models are compressible (shown in \cref{fig:fitting_random_labels} right). This is not just an empirical fact, but one that can be proven apriori: models which fit random labels cannot be compressed. While this result is a trivial consequence of complexity theory, we present an argument here for illustration.

\textbf{Almost all random datasets are incompressible}

When sampling labels uniformly at random, almost all datasets are not substantially compressible. Given a dataset $\mathcal{D} = \{x_i,y_i\}_{i=1}^n$ (where we are only considering the labels $y_i$, and conditioning on the inputs $x_i$), and denoting $|\mathcal{D}|$ as the length of the string of labels, the probability that a given dataset can be compressed to size $|\mathcal{D}|-c$ is less than $2^{-c+1}$. To see this, one must consider that there are only ${\sum_{i=0}^{|\mathcal{D}|-c} 2^i  \le 2^{|\mathcal{D}|-c+1}}$ programs of length $\le |\mathcal{D}|-c$ (fewer still when restricting to self delimiting programs), and there are $2^{|\mathcal{D}|}$ possible datasets. Therefore averaging over all randomly labeled datasets the fraction which are compressible to less than or equal to $|\mathcal{D}|-c$ bits is at most ${2^{|\mathcal{D}|-c+1}/2^{|\mathcal{D}|} = 2^{-c+1}}$.

\textbf{A compressible model which fits the data is a compression of the dataset}

Let prior $P$ that includes a specification of the model architecture, and the model $h$ which outputs probabilities for each of the outcomes: $p(y=k \mid x_i)=h(x_i)_k$. We can decompose the (prefix) Kolmogorov complexity of the dataset (given the prior) as 
\begin{equation}\label{eq:kolmogorov}
    K(\mathcal{D}\mid P) \le K(\mathcal{D} \mid h,P) + K(h \mid P).
\end{equation}

The term $K(\mathcal{D}\mid h,P)$ can be interpreted as a model fit term and upper bounded by the total negative log likelihood simply using the model probabilities as a distribution to encode the labels: ${K(\mathcal{D} \mid h,P) \le -\sum_i \log_2 h(x_i)_{y_i} +1 = \mathrm{NLL}(\mathcal{D} \mid h)+1}$.

Using the fact that almost all random datasets are incompressible, and choosing ${c=1+\log_2(1/\delta)}$, we have that with probability at least $1-\delta$ over all randomly sampled datasets ${K(\mathcal{D} \mid P)>|\mathcal{D}| -\log_2(1/\delta)-1}$.
Plugging into \cref{eq:kolmogorov} and rearranging, we have with probability $1-\delta$,
\begin{equation}
    K(h|P) \ge  \lvert{\mathcal{D}}\rvert - \mathrm{NLL}(\mathcal{D} \mid h)-\log_2(1/\delta) -2,
\end{equation}

In \cref{fig:model_width} we plot the quantity $K(h \mid P) + \mathrm{NLL}(\mathcal{D} \mid h)$ which represents the compressed size of the dataset achieved by our model (related to the minimum description length principle). We see that the value is considerably lower than the size of the dataset $|\mathcal{D}|$, emphasizing that real machine learning datasets such as CIFAR-10 have a very low Kolmogorov complexity and are very unlike those with random labels.
\section{Subspace Dimension Optimization and Hyperparameters in the Universal Prior} \label{subsec:prior}

The smaller the chosen intrinsic dimension $d$, the more similar $\theta$ is to the initialization $\theta_0$ in \cref{eq:ID}. Consequently, that value of $\theta$ is more likely under the universal prior given the shorter description length. Note that in this prior, we condition on the random seed used to generate $\theta_0$ and $P$.
As we optimize over different parameters such as the subspace dimension $d=1,..,D$, and possibly other hyperparameters such as the learning rate, or number of quantization levels $L$, we must encode these into our prior and thus pay a penalty for optimizing over them. We can accomodate this very simply by considering the hypothesis $h$ as not just specifying the weights, but also specifying these hyperparameters: $h = (\theta,d,L,\mathrm{lr})$, and therefore using the universal prior $P(h) = 2^{-K(h)}/Z$ we pay additional bits for each of these quantities:
$K(h) \le K(\theta\mid d,L)+K(d)+K(L) + K(\mathrm{lr})$. If we optimize over a fixed number $H$ of distinct values known in advance for a given hyperparameter such as $L$, then we can code $L$ using this information in only $\log_2(H)$ bits. In general, we can also bound the dimensionalities searched over by the maximum $D$ so that $K(d)\le \ceil{\log_2D}$ in any case.

\section{Transfer Learning Bounds}
\label{app-sec:transfer-learning-bounds}
We show the expanded results both with and without transfer learning in \autoref{table:additional-di-our-bounds}. When finetuning from ImageNet we use the larger EfficientNet-B0 models rather than the small convnet. Despite the fact that the model is significantly larger than the convnet or resnet models that we use to achieve the best bounds for from scratch training, the difference between the finetuned and pretrained models is highly compressible.

\section{Data Dependent Priors}
\label{app-sec:ddp}
\begin{figure}[!ht]
    \centering
    \begin{tabular}{cc}
       \includegraphics[width=.35\linewidth]{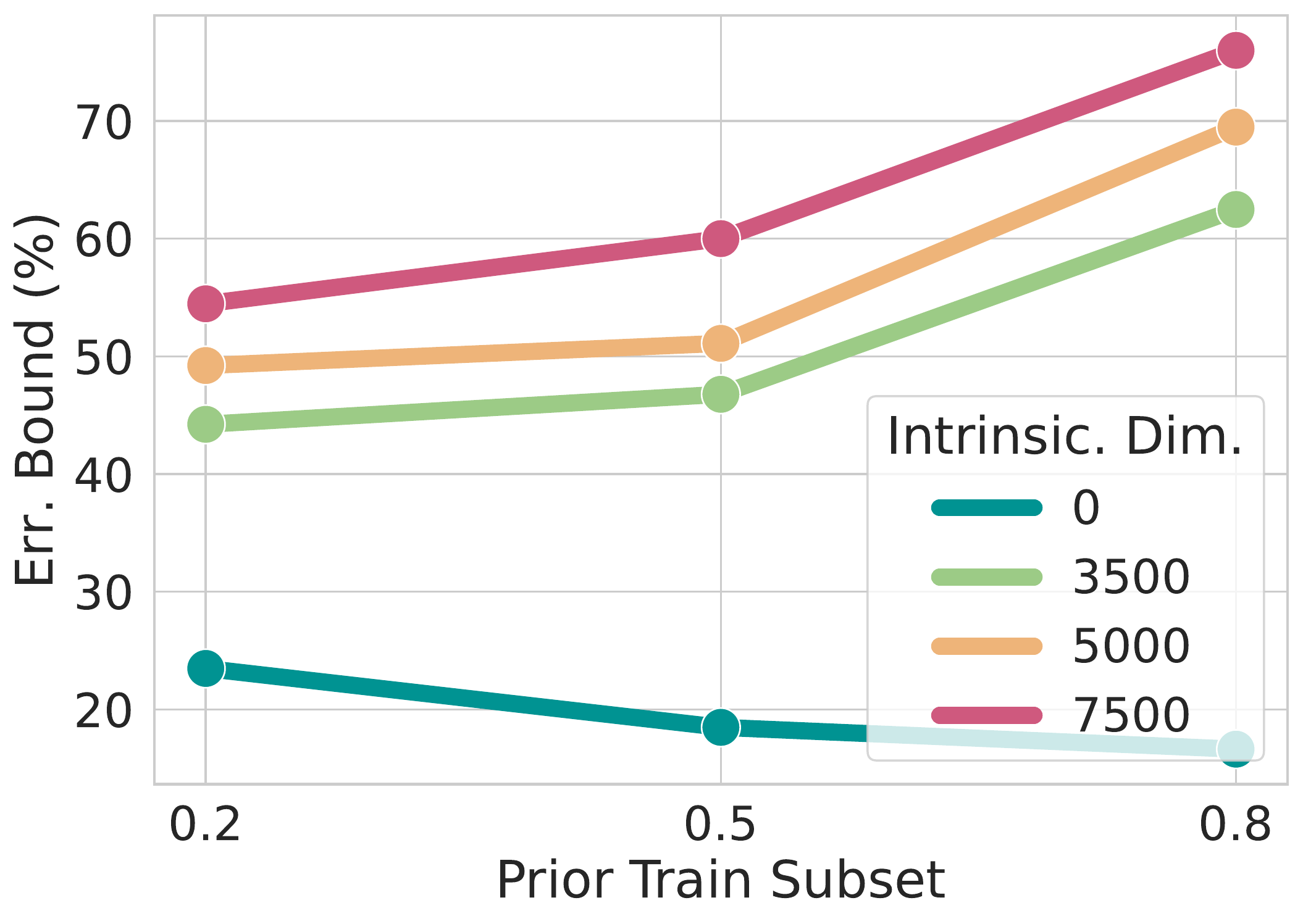}  &  \includegraphics[width=.35\linewidth]{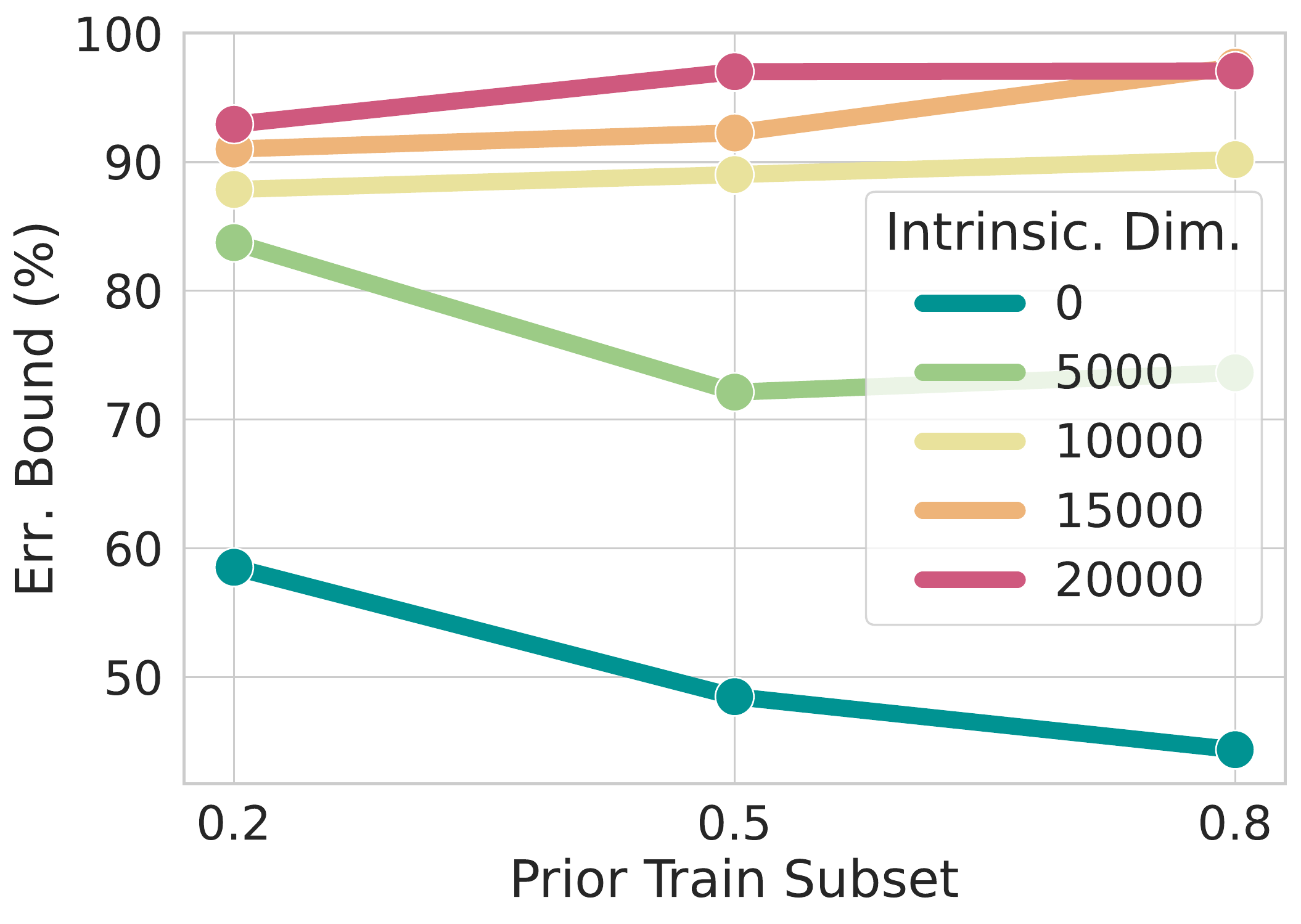} \\
       (a) CIFAR-10  & (b) CIFAR-100
    \end{tabular}
    \caption{ \textbf{Data-dependent bounds focus on fitting a good prior.}
    Our bounds using data dependent priors trained using varying fractions of the training dataset. We see that when using data dependent priors, lower intrinsic dimensionalities and lower KL models are favored by the bound.}
    \label{fig:data_dep_comparison}
\end{figure}

We observe that when using data dependent priors, our optimization over the subspace dimension (and the complexity of the model used to fit the data when measured against the prior) favors very low dimensions and low KL values which we show empirically in \autoref{fig:data_dep_comparison}. 
Indeed, a large fraction of the data fitting is moved into fitting a good prior, particularly when the dataset fraction used to train the prior is large. When the prior is already fitted on the data, the final solution can have a very low complexity with respect to that prior without affecting data fit, and is encouraged to do so.

\section{Experimental Details}
\label{app-sec:bound-hypers-optim}

In this section we provide experimental details to reproduce our results. 

\subsection{Model Training Details} \label{app-sec:model_training}
We use a standard small convolutional architecture for our experiments, which we find produces better bounds than its ResNet counterparts. The architecture is detailed in \cref{table:architecture}, and we use $k=16$ for experiments, but this value is ablated in \cref{fig:model_width}.

\textbf{Stochastic training:} All models were trained for $500$ epochs using the Adam optimizer with learning rate $0.001$, except for ImageNet which was trained for $80$ epochs with SGD using learning rate of $0.05$ and weight decay of $0.00002$. The model architectures for each dataset are listed below:
\begin{itemize}
    \item MNIST \citep{LeCun1998GradientbasedLA} ($+$ SVHN \citep{netzer2011reading} Transfer): LeNet-5 \citep{LeCun1998GradientbasedLA}.
    \item FashionMNIST \citep{xiao2017} ($+$ CIFAR-10 \citep{Krizhevsky2009LearningML} Transfer): ResNet20 \citep{He2016DeepRL}.
    \item SVHN: ConvNet (\cref{table:architecture}).
    \item SVHN $+$ ImageNet Transfer: EfficientNet-B0 \citep{Tan2019EfficientNetRM}.
    \item CIFAR-10: ConvNet.
    \item CIFAR-10 $+$ ImageNet Transfer: EfficientNet-B0 \citep{Tan2019EfficientNetRM}.
    \item CIFAR-100 \citep{Krizhevsky2009LearningML}: ConvNet.
    \item CIFAR-100 $+$ ImageNet \citep{deng2009imagenet} Transfer: EfficientNet-B0.
\end{itemize}

\begin{table}[!ht]
\centering
\caption{Simple convolutional architecture we use to compute our bounds.}
\label{table:architecture}
\begin{tabular}{ l l }
\noalign{\medskip}
\bf{ConvNet Architecture} \\
\hline\noalign{\smallskip}
Conv($3$,$k$), BN, ReLU\\
Conv($k$,$k$), BN, ReLU\\
Conv($k$,$2k$), BN, ReLU\\
MaxPool2d(2) \\
\hline
Conv($2k$,$2k$), BN, ReLU\\
Conv($2k$,$2k$), BN, ReLU\\
Conv($2k$,$2k$), BN, ReLU\\
MaxPool2d(2) \\
\hline
Conv($2k$,$2k$), BN, ReLU\\
Conv($2k$,$2k$), BN, ReLU\\
Conv($2k$,$2k$), BN, ReLU\\
GlobalAveragePool2d \\
Linear($2k$,$C$) \\
\hline\noalign{\smallskip}
\end{tabular}
\end{table}

\textbf{Full-batch training:} 
We train all models for $3000$ epochs, use learning rates equal to $0.1$ (MNIST $+$ LeNet-5 and CIFAR-10 $+$ ResNet-18) and $0.5$ (CIFAR-10 $+$ ConvNet), and a cosine learning rate scheduler that we warm-up for $10$ epochs. 
We also clip the full gradient to have an $L_2$-norm of at most $0.25$ before performing parameter updates in each epoch \citep{geiping2022stochastic}.

\textbf{Transfer Learning} All previous training details remain the same, except that $\theta_0$ from \cref{eq:ID} is initialized from a pre-trained checkpoint instead of a random initialization. As typically done in literature, the final classification layer is replaced with a randomly initialized fully-connected layer to account for the number of classes in the downstream task.

\subsection{Bound Hyperparameter Optimization}

As explained in \cref{subsec:prior}, we optimize the bound hyperparameters by considering that the hypothesis of interest $h$  specifies the hyperparameters in addition to the weights. 
Therefore, we pay bits back for the combination of hyperparemeters that we select. 
For example, if we are doing a grid search over $2$ values of the quantization-aware training learning rate, $2$ values of the intrinsic dimensionality values, $2$ values of the quantization levels, and use k-means by default, then the number of bits that we pay is $\log_2(2 \times 2 \times 2) = 3$ bits.

\textbf{Optimizing PAC-Bayes bounds for data-independent priors: } Our PAC-Bayesian subspace compression bounds for data-independent priors have $4$ hyperparameters that we list here-under alongside the possible values that we consider for each hyperparameter: 

\begin{itemize}
    \item The learning rate for the quantization-aware training, possible values: $\{0.001, 0.003, 0.005,  \allowbreak 0.0001\}$.
    \item The intrinsic dimensionality, possible values: $\{0, 1000, 2500, 3000, 3500, 4000, 5000, 7500, \allowbreak 8000, 10000, 12000, 15000, 20000, 25000, 50000, 100000, 250000, 500000\}$, except for the ImageNet transfer learning which was conducted on the more limited range: $\{500,1000,2000,3000,4000,6000,8000\}$
    \item The number of quantization levels, possible values: $\{0, 7, 11, 30, 50\}$.
    \item The quantization initialization, possible values: $\{\text{uniform}, \text{k-means}\}$. 
\end{itemize}

Note that we only use a subset of these hyperparameter values for some datasets, depending on the dataset size and other considerations. 
For all bound computations, we use arithmetic encoding and $30$ epochs of quantization-aware training.

In \cref{table:data-indep-bounds-hypers}, we summarize the hyperparameters corresponding to the data-independent bounds that we report in  \cref{table:our-bounds}.

\begin{table}[ht]
  \caption{Hyperparameters corresponding to our PAC-Bayesian Subspace Compression Bounds reported in \cref{table:our-bounds} as well as SVHN and ImageNet to SVHN transfer learning with \textbf{data-independent priors}. All bound results are obtained with $95\%$ confidence, i.e. $\delta = .05$. 
  }
  \label{table:data-indep-bounds-hypers}
  \centering
\begin{adjustbox}{width=\linewidth}
\begin{tabular}{l|ccccc}
\hline
 & Err. Bound (\%) & Quant. Learning Rate & Intrinsic Dimensionality & Levels  & Quant. Init.\\
  \hline 
 MNIST  & $\mathbf{11.6}$ & $0.005$ & $1000$ & $7$ & Uniform \\
 \phantom{+} + SVHN Transfer & $\mathbf{9.0}$ & $0.005$ & $1000$ & $7$ &  Uniform \\
 \hline
 FashionMNIST  & $\mathbf{32.8}$ & $0.005$ & $2500$ & $7$ & Uniform \\
 \phantom{+} + CIFAR-10 Transfer & $\mathbf{28.2}$ & $0.005$ & $2500$ & $7$ & Uniform \\
 \hline
  SVHN & 
  $\mathbf{36.1}$ & $0.0001$ & $3500$ & $11$ & Uniform \\
 \phantom{+} + ImageNet Transfer & 
 $\mathbf{29.1}$ & $0.003$ & $4000$ & $7$ & Uniform \\
 \hline
 CIFAR-10 & $\mathbf{58.2}$ & $0.0001$ & $3500$ & $7$ & k-Means \\
 \phantom{+} + ImageNet Transfer& $\mathbf{35.1}$ & $0.003$ & $3000$ &7 & Uniform\\
 \hline
 CIFAR-100  & $\mathbf{94.6}$ & $0.0001$ & $10000$ & $11$ & k-Means \\
 \phantom{+} + ImageNet Transfer &$\mathbf{81.3}$ & $0.003$ & $8000$ & 7& Uniform\\
 \hline
\end{tabular}
\end{adjustbox}
\end{table}

\textbf{Optimizing PAC-Bayes bounds for data-dependent priors:} In addition to the hyperparameters listed above, we also tune the hyperparameter corresponding to the subset of the training dataset that we use to train the prior on. 
We consider the following values for the subset of the training dataset: $\{20\%, 50\%, 80\%\}$.

In \cref{table:data-dep-bounds-hypers}, we summarize the the hyperparameters corresponding to the data-dependent bounds that we report in  \cref{table:our-bounds}. 
The best bounds are obtained for intrinsic dimensionality equal to $0$, therefore no quantization is performed. 

\begin{table}[ht]
  \caption{Hyperparameters corresponding to our PAC-Bayes bounds reported in \cref{table:our-bounds} as well as SVHN and ImageNet with \textbf{data-dependent priors}. The best bounds are obtained for intrinsic dimensionality equal to $0$, therefore no quantization is performed. All bound results are obtained with $95\%$ confidence, i.e. $\delta = .05$.}
  \label{table:data-dep-bounds-hypers}
  \centering
\begin{adjustbox}{width=0.6\linewidth}
\begin{tabular}{l|cc}
\hline
 & Err. Bound (\%) & Training Subset (\%) \\
  \hline 
 MNIST  & $\mathbf{1.4}$ &  $50$ \\
 \hline
 FashionMNIST  & $\mathbf{10.1}$ & $80$ \\
 \hline
 SVHN  & $\mathbf{8.7}$ & $50$ \\
 \hline
 CIFAR-10 & $ \mathbf{16.6}$ & $80$ \\
 \hline
 CIFAR-100  & $\mathbf{44.4}$ & $80$ \\
 \hline
 ImageNet  & $\mathbf{40.9}$ & $50$ \\
 \hline
\end{tabular}
\end{adjustbox}
\end{table}

\subsection{Computational Infrastructure \& Resources}

Our computational hardware involved a mix of NVIDIA GeForce RTX 2080 Ti (12GB), NVIDIA TITAN RTX (24GB), NVIDIA V100 (32GB), and NVIDIA RTX8000 (48GB). The experiments were managed via W\&B \citep{wandb}. The total computational cost of all experiments (including the ones that do not appear in this work) amounts to $\approx 8000$ GPU hours.

\subsection{Breaking Data and Model Structure Experiment}\label{sec-app:breaking_structure}
\begin{figure}[!ht]
\centering
    \begin{tabular}{cc}
    \includegraphics[width=.4\linewidth]{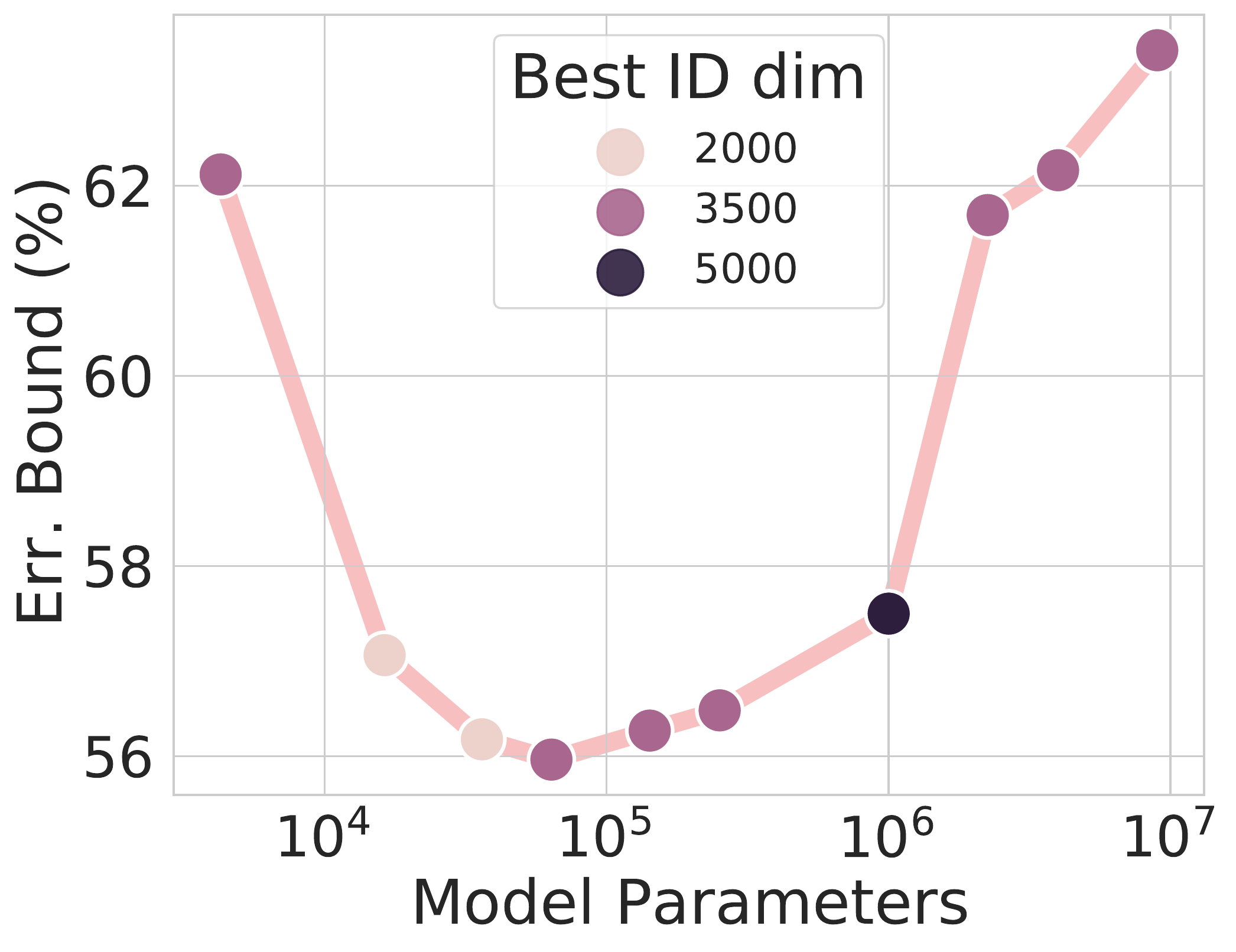} &
    \includegraphics[width=.43\linewidth]{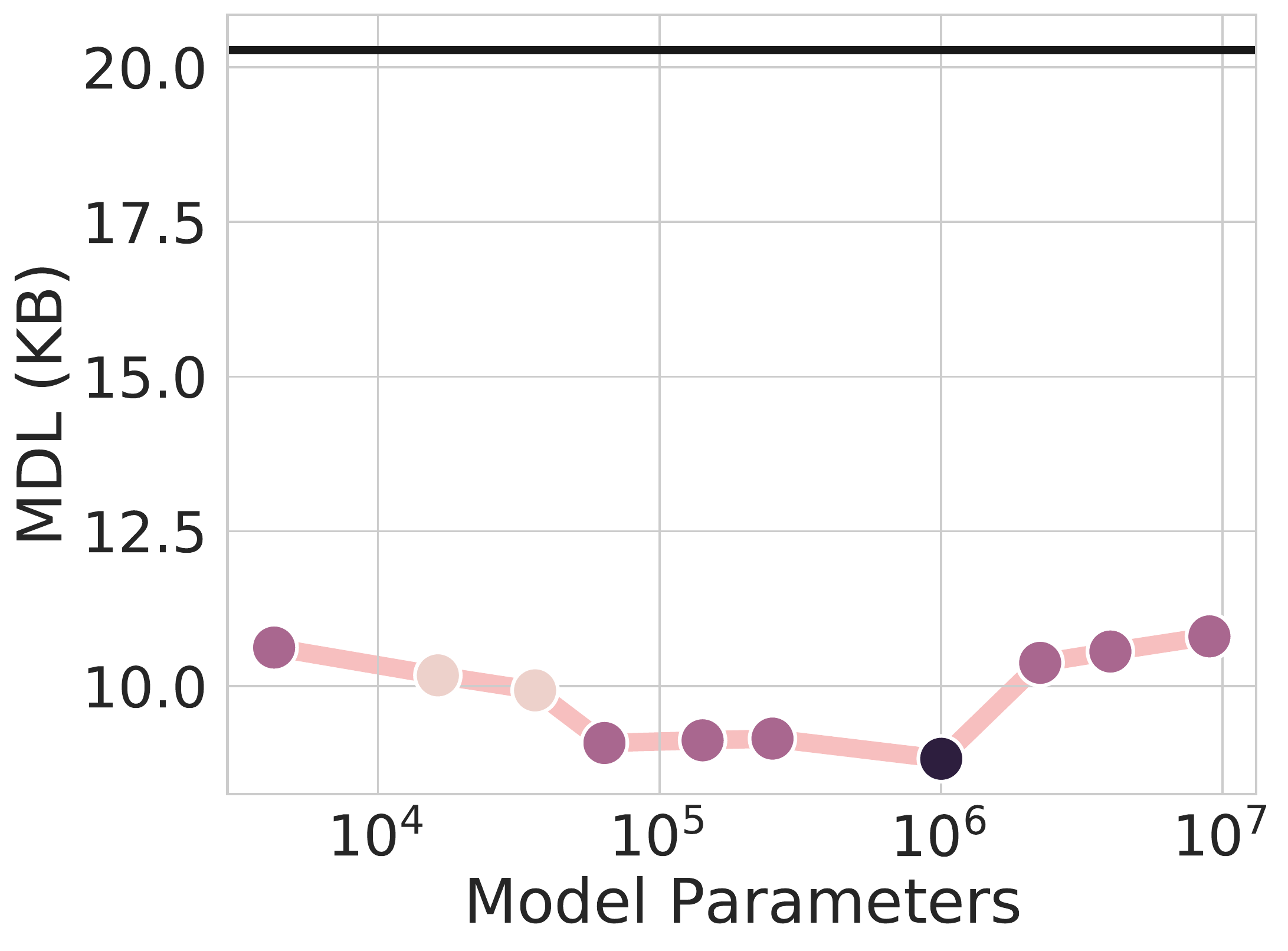}
    \end{tabular}
   \caption{ \textbf{Model size, compressibility, and MDL}.
   \textbf{Left:} Generalization error bound as a function of model size on the CIFAR10 dataset. The ID subspace dimension that achieves the best bound is shown by the color. In terms of our bound computation, the optimal number of parameters of the network is only slightly above the number of data points. 
   \textbf{Right:} The total compressed size ($K(h|P)+\mathrm{NLL}$) of the training dataset using our model as a compression scheme. While the raw labels have size $20.3$KB (shown by the black line), the best model compresses the labels down to $8.6$KB. Curiously, the compressed dataset size and hence the MDL principle favors larger models than our generalization bounds.}
    \label{fig:model_width}
\end{figure}

In this experiment we compared our generalization bounds derived for training convolutional networks and MLPs on standard CIFAR10, as well as when data structure is broken by shuffling the pixels or shuffling the labels.
We trained for 100 standard epochs with batch size 128 and then another 50 epochs of quantization aware training in all cases. We use 7 quantization levels and uniform quantization initialization for all to simplify. When comparing against an MLP, we use a 3 hidden layer MLP with ReLU nonlinearities, and we feed in the images by flattening them into $3\times 32 \times 32$ sized vectors. We use $150$ hidden units in the intermediate layers of the MLP and choose $k=46$ in the simple convolutional architecture described in \cref{table:architecture} so as to match the parameter count (though slightly smaller models perform slightly better as ablated in \cref{fig:model_width}).

\section{Equivariance}\label{sec:app_equivariance}
We conduct a simple experiment to evaluate the extent to which model equivariance has on the compressibility of deep learning models and the tightness of our generalization bounds. We use the rotationally equivariant $C_8$ WideResNet model from \citet{weiler2019general} which has an $8$-fold rotational symmetry, and we also use a non equivariant version of this model. The equivariant model has a depth of $10$ and a widen factor of $4$ yielding $1.451$M parameters. We control for the number of parameters by adjusting the widen factor of the non equivariant model to $4.67$ yielding $1.447$M parameters.

\begin{figure}[!ht]
    \centering
    \begin{tabular}{cc}
        \includegraphics[width=.4\linewidth]{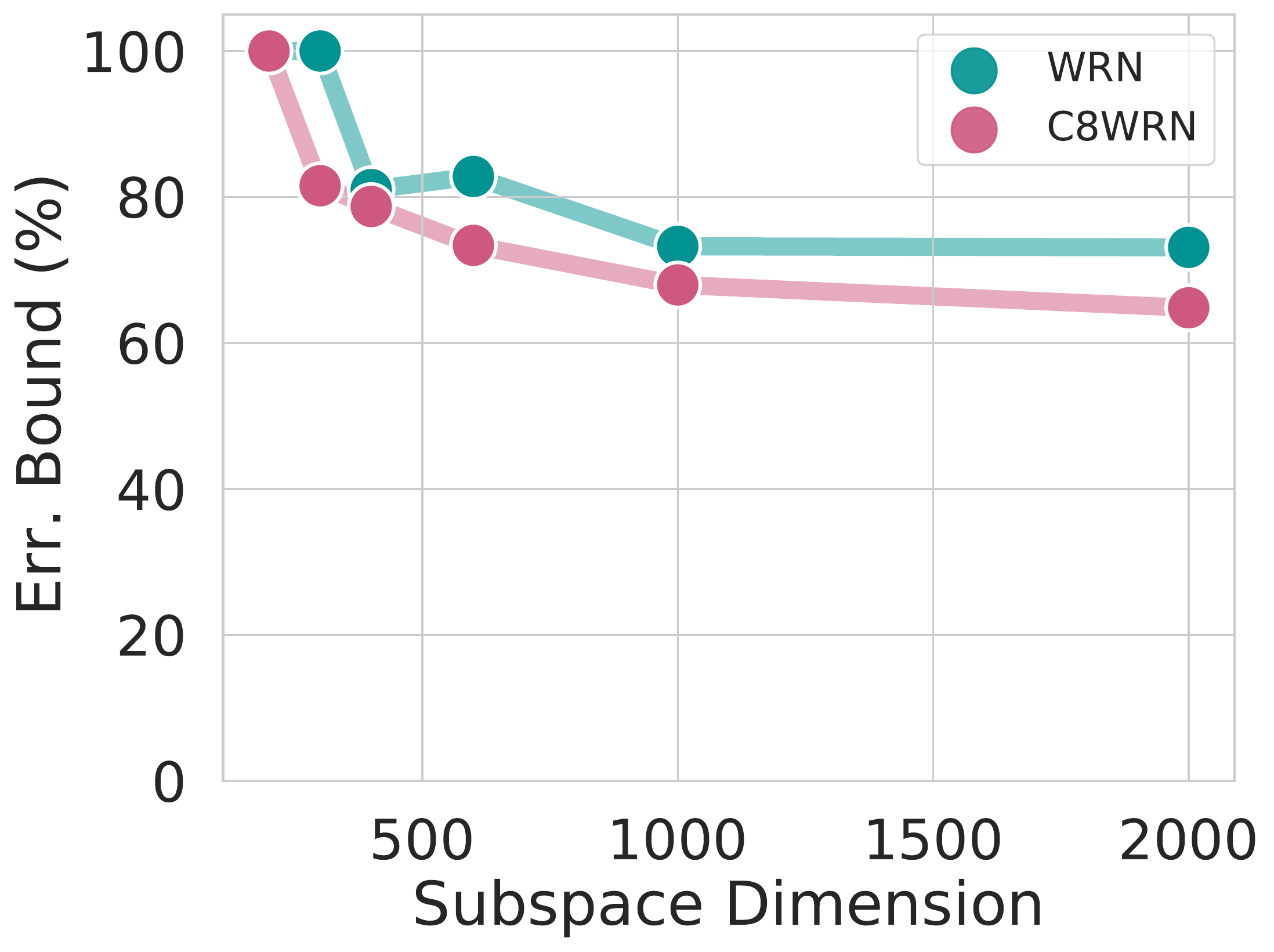}&
        \includegraphics[width=.4\linewidth]{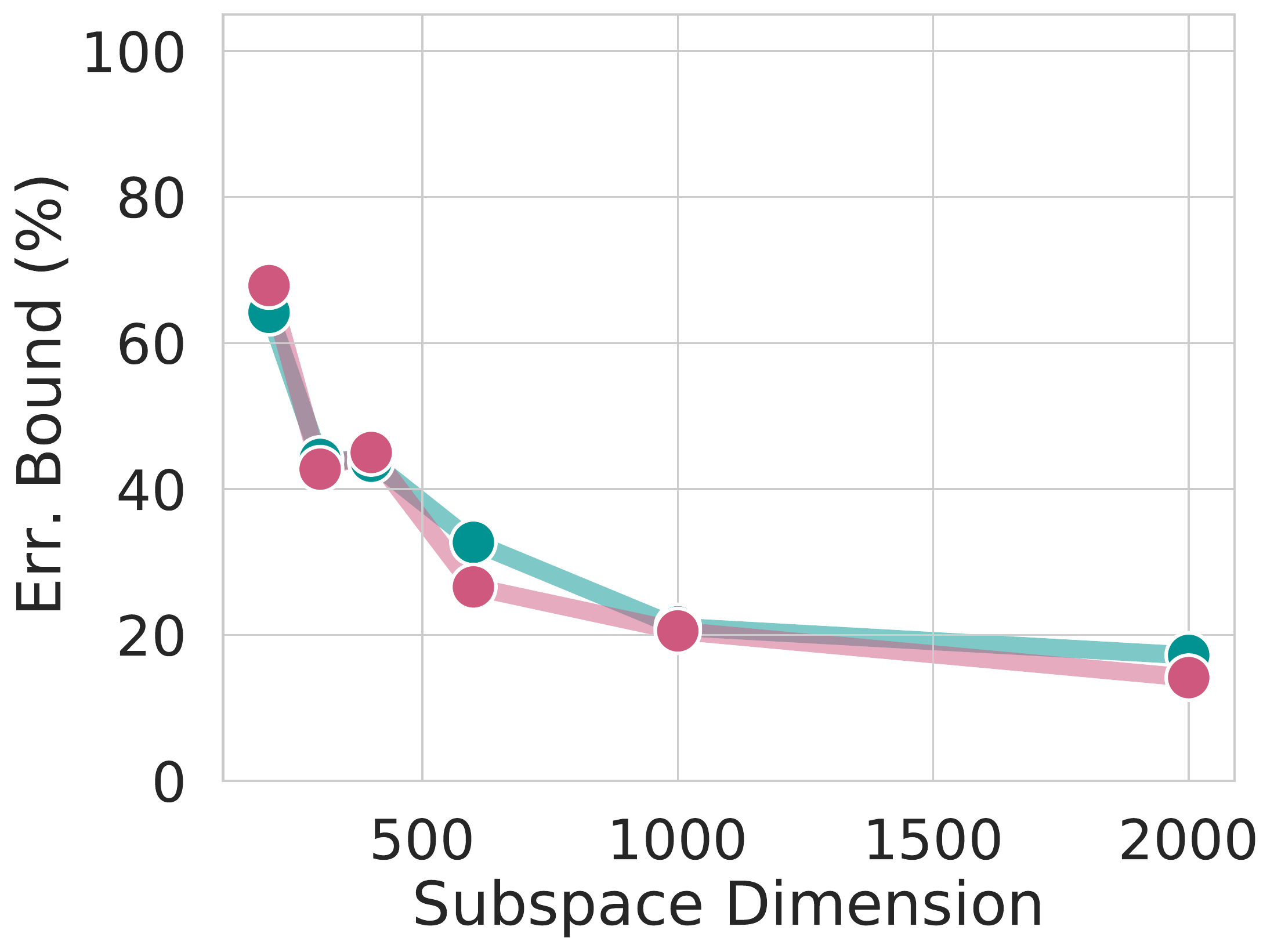}\\
        (a) RotMNIST (12k labels) & (b) MNIST (60k labels)  \\
    \end{tabular}
    \caption{ \textbf{Rotationally-equivariant models provably generalize better on rotationally-equivariant data.}
    Comparison of rotationally equivariant $C_8$ WideResNet vs ordinary WideResNet with the same number of parameters on (a) the rotationally equivariant RotMNIST dataset \citep{larochelle2007empirical} and (b) the ordinary MNIST dataset. Both models are capable of fitting the data, but the equivariant model yields a more compressible solution when fitting the rotationally equivariant data than the non equivariant model, and hence yields a better generalization bound. (Note the difference in dataset size, RotMNIST has only 12K data points unlike MNIST)}
    \label{fig:equivariance}
\end{figure}

We evaluate these models both on MNIST and the RotMNIST dataset \citep{larochelle2007empirical} consisting of 12K training examples of rotated MNIST digits. As shown in \autoref{fig:equivariance} (a), when paired with the rotationally symmetric RotMNIST dataset, the rotationally equivariant model achieves better bounds and is more compressible than it's non equivariant counterpart despite having the same number of parameters. However, when this symmetry of the dataset is removed by considering standard MNIST, we see that the benefits of equivariance to the generalization bound and compressibility vs the WRN model dissapear.

\section{Full-Batch vs. Stochastic Training (SGD)}
\label{app-sec:stochasticity}

To further expand on the results that we present in \cref{sec:understanding-gen}, we study the impact of hyperparameters, namely the weight decay and the architecture choice, on the bounds obtained through full-batch (F-B) training. 
\cref{table:bounds-full-batch} summarizes these results and we provide the training and bound computation details in \cref{app-sec:bound-hypers-optim}.
Our PAC-Bayes subspace compression bounds provide similar theoretical guarantees for both full-batch and stochastic training, suggesting that the implicit biases of SGD are not necessary to guarantee good generalization. 
Moreover, we see that the results are consistent for different configurations, which result in comparable bounds overall. 

\textbf{Transfer learning using full-batch training:} We perform full batch training for transfer learning from SVHN to MNIST using LeNet-5 and the same experimental setup described in \cref{app-sec:bound-hypers-optim}.
Our best PAC-Bayes subspace compression bounds for SVHN to MNIST transfer are $8.7\%$ and $9.0\%$ for full-batch and SGD training, respectively. 
This finding provides further evidence that good generalization of neural networks, and  the success of transfer learning in particular, does not necessarily require stochasticity or additional flatness-inducing procedures to be achieved. 

Finally, we note that we optimize over the same set of hyperparameters for the bound computation for both full-batch and stochastic training.  

\begin{table}[!ht]
  \caption{Our PAC-Bayes subspace compression bounds obtained through full-batch (F-B) training for different configurations and datasets.}
  \label{table:bounds-full-batch}
  \centering
\begin{adjustbox}{width=0.9\linewidth}
\begin{tabular}{l|ccccc}
\hline
 Dataset & Architecture & Stochastic Err. Bound (\%) & F-B Weight Decay  & F-B Err. Bound (\%) \\
  \hline 
 MNIST  & \specialcell{LeNet-5} & \specialcell{$11.6$} & \specialcell{$0.01$ \\ $0.001$ \\ $0.005$ \\ $0.0001$} & \specialcell{$12.5$ \\ \textbf{$\mathbf{11.2}$} \\ $12.0$ \\ $11.7$} \\
 \hline
  CIFAR-10  & \specialcell{ResNet-18} & \specialcell{$74.7$} & \specialcell{$0.01$ \\ $0.001$ \\ $0.005$ \\ $0.0001$} & \specialcell{$77.8$ \\ $ 76.3$ \\ $76.1$ \\ $\mathbf{75.3}$} \\
 \hline
  CIFAR-10  & \specialcell{ConvNet} & \specialcell{$58.2$} & \specialcell{$0.01$ \\ $0.001$ \\ $0.0001$} & \specialcell{ $65.8$ \\ $63.6$ \\ $\mathbf{61.4}$} \\
 \hline
\end{tabular}
\end{adjustbox}
\end{table}

\section{Model Size vs. Compressibility} \label{sec:modelsize}
We perform an ablation to determine how the size of the model affects our generalization bounds. Using the fixed model architecture \autoref{table:architecture}, we vary the width $k$ from $4$ to $192$. Using our subspace compression scheme, we find that the compression ratio of the model does increase with model size, however the total compressed size still increases slowly making our bounds less strong for larger models. For this paper, we find the sweet spot $k=16$ is just above the point with equal number of parameters and data points.

We note that this finding leaves room for an improved compression scheme and generalization bounds which are able to explain why even larger models still generalize better. Curiously, when plotting to the total compressed dataset size ($K(h|P)+\mathrm{NLL}$) using the model as a compression scheme, we find that the MDL principle which favors shorter description lengths of the data actually prefers larger models than our PAC-Bayes generalization bounds selects.

\section{Double Descent}
\label{app-sec:double-descent}
\begin{figure}[!ht]
\centering
    \begin{tabular}{cc}
    \includegraphics[width=.4\linewidth]{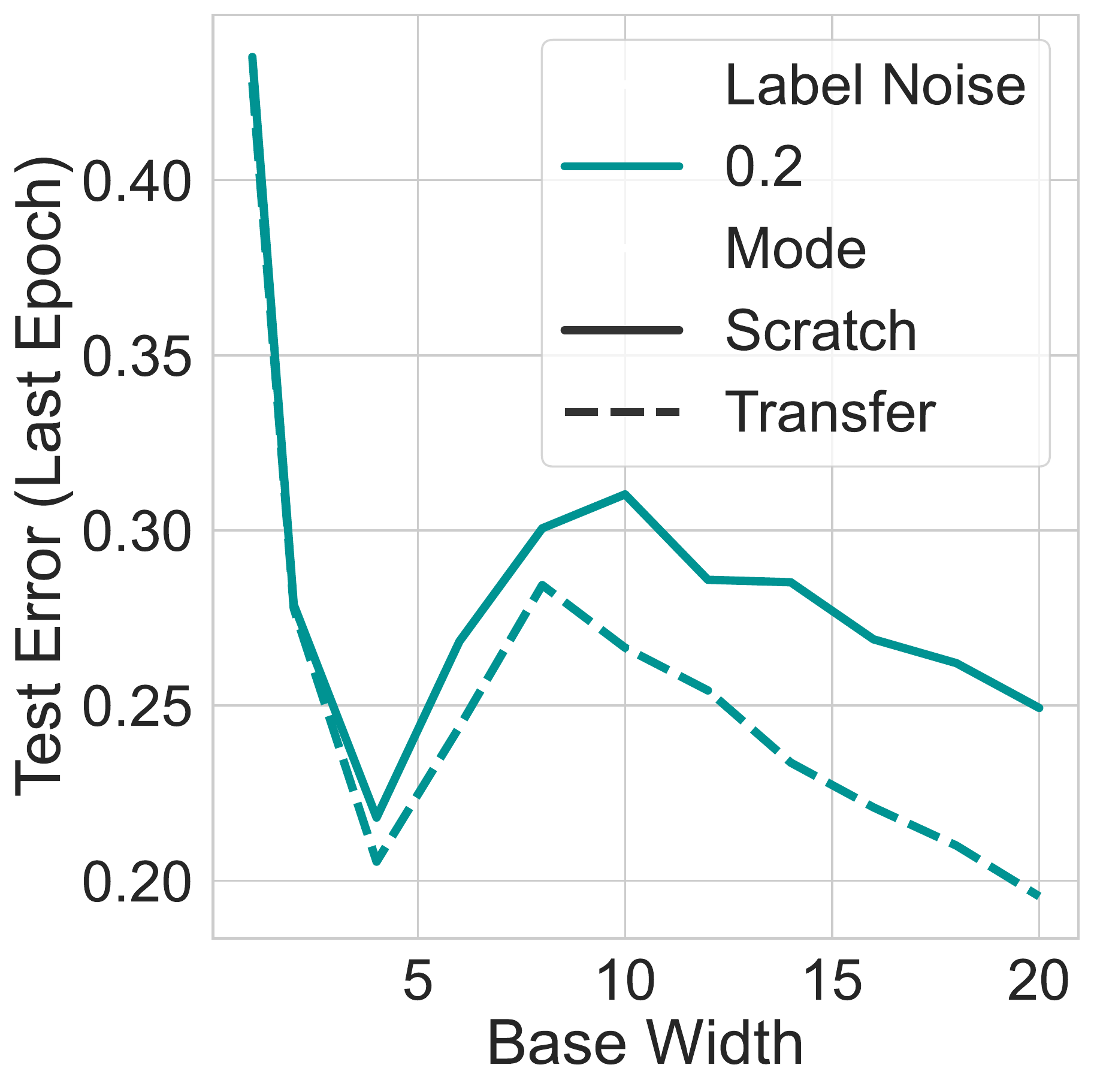} &
    \includegraphics[width=.4\linewidth]{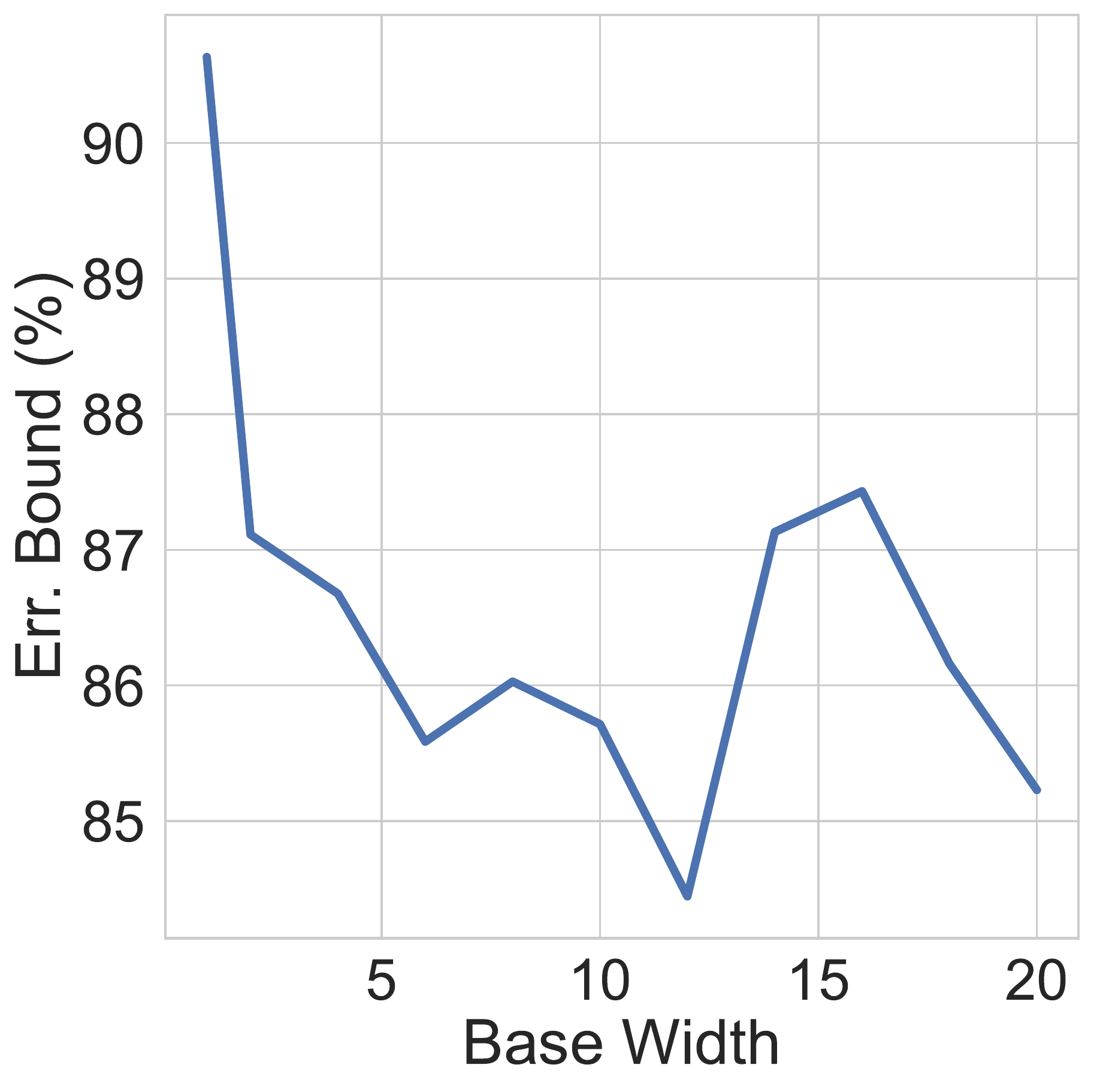}
    \end{tabular}
   \caption{ \textbf{Our bounds display a double descent as we increase the width.}
   \textbf{Left:} Double descent (in terms of the test error of the last epoch model) observed when varying the width of a ResNets-18 model to fit the CIFAR-10 dataset with label noise equal to $0.2$. 
   \textbf{Right:} Our bounds showing a similar \textit{double descent} behaviour where the bound starts to worsen only to become better again at a later width. Here we can fix the intrinsic dimensionality to be equal to $35000$ and we choose the best subspace compression bound for each base width.}
    \label{fig:dd-bounds}
\end{figure}

Under select conditions, we are able to reproduce the double descent phenomenon in our generalization bounds. In \cref{fig:dd-bounds} (right), we show that our bound exhibits a double descent similar to what we see in terms of the test error \cref{fig:dd-bounds} (left).  The results we show in \cref{fig:dd-bounds} (right) are obtained for a fixed intrinsic dimensionality of $35000$, but we observed that this middle descent consistently appears in our bounds plots for a given (fixed) intrinsic dimensionality where we select the best bound for each base width. However, we expect that extending the plot out to larger model widths the bound gets worse again as explained in \cref{sec:modelsize}. 

\section{PAC-Bayes Bounds} \label{sec:catoni}%\label{sec:catoni}

\subsection{Catoni PAC-Bayes Bound} 
In our case, since neural networks achieve low training error, we focus on a bound like 
\citet{catoni2007pacbayes} which becomes tighter when $\mathbb{KL}\left(Q, P\right)$ is large. This is the same bound used in \citet{zhou2019nonvacuous}.
\begin{theorem}[\citet{catoni2007pacbayes}]
\label{th:catoni}
Given a $0$-$1$ loss $\ell$, a fixed $\alpha > 1$ and a confidence level $\delta \in \left(0, 1\right)$ then
\begin{equation*}
    \begin{split}
    \underset{\mathbb{E}}{\theta \sim Q}[R\left(f_{\theta}\right)]
      \leq
      \inf_{\lambda > 1} \Phi^{-1}_{\lambda / N}\left[
        \underset{\mathbb{E}}{\theta \sim Q}[
          \hat{R}\left(f_{\theta}\right)] + 
          \frac{\alpha}{\lambda} \left[
            \mathbb{KL}\left(Q, P\right) + \log \frac{1}{\delta} + 2 \log \left(\frac{\log\left(\alpha^{2} \lambda\right)}{\log \alpha}\right)
            \right]
        \right]
    \end{split}
    \label{eq:pac_catoni}
\end{equation*}
holds with probability higher than $1 - \delta$ and where
\begin{equation*}
    \begin{split}
      \Phi^{-1}_{\gamma}\left(x\right) = \frac{1 - e^{\gamma x}}{1 - e^{\gamma}}.
    \end{split}
\end{equation*}
\end{theorem}

\subsection{Variable Length Encoding and Robustness Adjustment}
In \citet{zhou2019nonvacuous}, the authors assume a fixed length encoding for the weights. Given that the distribution over quantization levels is highly nonuniform, using a variable length encoding (such as Huffman encoding or
arithmetic encoding) can represent the same information using fewer bits. While this choice gives significant benefits, it means that we cannot immediately make use of robustness adjustment from \citet{zhou2019nonvacuous}, where the robustness adjustment comes from considering neighboring models
that result from perturbing the weights slightly.

Revisiting the prior derivation in \citet{zhou2019nonvacuous}, we show why the method used for bounding the KL does not transfer over to variable length encodings.
In \citet{zhou2019nonvacuous}, the prior used is
\begin{equation*}
    \begin{split}
        P = 
        \tfrac{1}{Z}\sum_{S, Q, C}^{} 2^{-(\left|S\right| + \left|C\right| + d \lceil \log L \rceil)} \mathcal{N}\left(\hat{w}\left(S, Q, C\right), \tau^{2}\right)
    \end{split}
\end{equation*}
where $S$ denotes the encoding of the position of the pruned weights, $C$ denotes the codebook, $Q$ the codebook value that the weight take, $d$ the number of 
nonzero weights, $L$ the number of clusters, $\hat{w}$ the quantized weight and $\tau^{2}$ the prior variance.
Note that $\hat{w}$ changes depending on $S, Q, C$ and also note that the fixed-length encoding can be seen in 
how we sum over $d \lceil \log L \rceil $ options.
This prior is a mixture of Gaussians centered at the quantized values.
With this choice of prior \citet{zhou2019nonvacuous} and setting the posterior to be also Gaussian centered at a quantized value, one can upper bound the KL with a computationally tractable term involving the sum over dimensions. 
Crucially, for their decomposition they use the fact that the size of the encoding $|Q|$ is $d\ceil{\log L}$, which is independent of the coding $Q$ and only dependent on the codebook $C$. Therefore they are able to upper bound the KL. 
\begin{equation*}
    \begin{split}
      \mathbb{KL}\left(\mathcal{N}\left(\hat{w}, \sigma^{2} I_d\right), \sum_{Q}^{} \mathcal{N}(\hat{w}(\hat{S}, Q, \hat{C}), \tau^2)\right)
      =
      \sum_{i=1}^{d} \mathbb{KL}\left(\mathcal{N}\left(\hat{w}_{i}, \sigma^{2}\right), 
      \sum_{j=1}^{L} \mathcal{N}\left(\hat{w}_{j}, \tau^{2}\right)\right)
    \end{split}
\end{equation*}
due to the independence of the fixed length encoding to that of the values that each quantized value takes, see appendix in \citet{zhou2019nonvacuous}.
This independence is broken for variable length encoding as the cluster centers and the values that each weight can take are interlinked.
Thus, we cannot express and satisfactorily approximate the first high-dimensional KL term as a sum of one-dimensional elements that can be estimated through quadrature or Monte Carlo.

\section{Licences}
\label{sec:licences}

MNIST \footnote{\url{http://yann.lecun.com/exdb/mnist/}} \citep{LeCun1998GradientbasedLA} is made available under the terms of the Creative Commons Attribution-Share Alike 3.0 license. FashionMNIST \footnote{\url{https://github.com/zalandoresearch/fashion-mnist}} \citep{xiao2017} is available under the MIT license. CIFAR-10 and CIFAR-100 \citep{Krizhevsky2009LearningML} are made available under the MIT license. ImageNet \footnote{\url{https://www.image-net.org/download}} \citep{deng2009imagenet} is the copyright of Stanford Vision Lab, Stanford University, Princeton University. SVHN \footnote{\url{http://ufldl.stanford.edu/housenumbers/}} \citep{netzer2011reading} is released for non-commericial use only.

\end{document}